\documentclass{article}

\usepackage[nonatbib,preprint]{neurips_2020}

\usepackage[utf8]{inputenc} 
\usepackage[T1]{fontenc}    
\usepackage{hyperref}       
\usepackage{url}            
\usepackage{booktabs}       
\usepackage{amsfonts}       
\usepackage{nicefrac}       
\usepackage{microtype}      
\usepackage{stfloats}
\usepackage{subfigure}
\usepackage{graphicx}
\usepackage{subfigure}
\usepackage{booktabs} 
\usepackage{pgfplots}
\usepackage{selectp}
\usepackage{hyperref}
\usepackage{amsmath}
\usepackage{listings}
\usepackage{bbm}
\usepackage{amssymb} 
\usepackage{mathrsfs}
\usepackage{algorithm}
\usepackage{algorithmic}
\usepackage{amsthm}
\usepackage{textcomp}
\usepackage{gensymb}
\usepackage{graphicx}
\usepackage{color}
\usepackage{xcolor}
\usepackage{accents}
\usepackage{stackengine}
\newcommand\xrowht[2][0]{\addstackgap[.5\dimexpr#2\relax]{\vphantom{#1}}}

\newtheorem{Theorem}{Theorem}
\newtheorem{Theorem*}{Theorem}

\newtheorem{Claim*}[Theorem]{Claim}
\newtheorem{Corollary}[Theorem]{Corollary}

\newtheorem{CounterExample*}{$\overline{\hbox{\bf Example}}$}

\newtheorem{Example*}[Theorem]{Example}

\newtheorem{Intuition*}[Theorem]{Intuition}
\newtheorem{Joke*}[Theorem]{Joke}
\newtheorem{Lemma}[Theorem]{Lemma}
\newtheorem{Lemma*}[Theorem]{Lemma}
\newtheorem{Open problem}[Theorem]{Open problem}

\newtheorem{Question*}[Theorem]{Question}

\def \bSubexa    {\begin{subexa}}

\newcommand{\ignore}[1]{}

\newcommand{\EE}{\mathbb{E}}
\newcommand{\CC}{\mathbb{C}}

\newcommand{\RR}{\mathbb{R}}
\newcommand{\ZZ}{\mathbb{Z}}

\newcommand{\reals}{\RR}

\def \cA     {{\cal A}}

\def \cC     {{\cal C}}
\def \cD     {{\cal D}}

\def \cN     {{\cal N}}
\def \cO     {{\cal O}}
\def \cP     {{\cal P}}
\def \cQ     {{\cal Q}}
\def \cR     {{\cal R}}

\def \upto  {{,}\ldots{,}}

\newcommand{\ed}{\stackrel{\mathrm{def}}{=}}

\newcommand{\bi}{\begin{itemize}}
\newcommand{\ei}{\end{itemize}}

\def\orpro{\mathop{\mathchoice
   {\vee\kern-.49em\raise.7ex\hbox{$\cdot$}\kern.4em}
   {\vee\kern-.45em\raise.63ex\hbox{$\cdot$}\kern.2em}
   {\vee\kern-.4em\raise.3ex\hbox{$\cdot$}\kern.1em}
   {\vee\kern-.35em\raise2.2ex\hbox{$\cdot$}\kern.1em}}\limits}

\def\andpro{\mathop{\mathchoice
 {\wedge\kern-.46em\lower.69ex\hbox{$\cdot$}\kern.3em}
 {\wedge\kern-.46em\lower.58ex\hbox{$\cdot$}\kern.25em}
 {\wedge\kern-.38em\lower.5ex\hbox{$\cdot$}\kern.1em}
 {\wedge\kern-.3em\lower.5ex\hbox{$\cdot$}\kern.1em}}\limits}

\def\simge{\mathrel{%
   \rlap{\raise 0.511ex \hbox{$>$}}{\lower 0.511ex \hbox{$\sim$}}}}

\def\simle{\mathrel{
   \rlap{\raise 0.511ex \hbox{$<$}}{\lower 0.511ex \hbox{$\sim$}}}}

\newcommand{\OPT}{\text{OPT}}
\newcommand{\Beta}{\text{Beta}}

\newcommand{\barI}{\bar{I}}

\newcommand{\intJ}{J}

\newcommand{\barIp}[1]{\barI_{#1}}

\newcommand{\barJ}{\bar{\intJ}}

\newcommand{\intp}{P}

\newcommand{\hatcump}{r}
\newcommand{\hatp}{q}
\newcommand{\hatpp}{p}
\newcommand{\hats}{s}

\newcommand{\barhatp}{\bar{\hatp}}
\newcommand{\barhatpp}{\bar{\hatpp}}
\newcommand{\barhats}{\bar{\hats}}
\newcommand{\barp}{\bar{p}}
\newcommand{\baru}[1]{(1/{#1}\upto1/{#1} )}

\newcommand{\hatf}{\hat{f}}

\newcommand{\starf}{f^{\star}}

\newcommand{\barn}{\bar{n}}
\newcommand{\barm}{\bar{m}}
\newcommand{\poly}{h}

\newcommand{\del}{\Lambda}
\newcommand{\thr}{\lambda}
\newcommand{\lemerge}{\mu}

\newcommand{\ordst}[2]{{#1}_{(#2)}}

\newcommand{\lone}[1]{\|#1\|_{1}}
\newcommand{\loneint}[2]{{\|#2\|}_{#1}}

\newlength{\dhatheight}

\newcommand{\EQ}{\mathrm{INT}}
\newcommand{\STITCH}{\mathrm{MERGE}}
\newcommand{\MERGE}{\mathrm{COMP}}
\newcommand{\SURF}{\mathrm{SURF}}
\newcommand{\ADLS}{\mathrm{ADLS}}

\let\epsilon\relax
\newcommand{\epsilon}{\varepsilon} 

\pgfplotsset{compat=1.17}
\DeclareMathOperator*{\E}{\EE}

\title{SURF: A Simple, Universal, Robust, Fast\\
	Distribution Learning Algorithm}

\author{%
  Yi Hao, Ayush Jain, Alon Orlitsky, Vaishakh Ravindrakumar\\
  Dept. of Electrical and Computer Engineering\\
  University of California, San Diego\\
  \texttt{\{yih179, ayjain, aorlitsky, varavind\}@eng.ucsd.edu} 
}

\begin{document}

\maketitle

\begin{abstract}
	\ignore{-.5em}
\vspace{-1em}
Sample- and computationally-efficient distribution estimation is a fundamental tenet in statistics and machine learning. We present $\SURF$, an algorithm for approximating distributions by piecewise polynomials. $\SURF$ is:
simple, replacing prior complex optimization techniques by straight-forward {empirical probability} approximation of each potential polynomial piece {through simple empirical-probability interpolation}, and using plain divide-and-conquer to merge the pieces; universal, as well-known  polynomial-approximation results imply that it accurately approximates a large class of common distributions; 
robust to distribution mis-specification as for any degree $d \le 8$, it estimates any distribution to an $\ell_1$ distance $< 3$ times that of the nearest degree-$d$ piecewise polynomial, improving known factor upper bounds of 3 for single polynomials and 15 for polynomials with arbitrarily many pieces;
fast, using optimal sample complexity, running in near sample-linear time, and if given sorted samples it may be parallelized to run in sub-linear time.
In experiments, $\SURF$ outperforms state-of-the art algorithms. 
\vspace{-1em}
	\ignore{-1.75em}
\end{abstract}
\section{Introduction}
\vspace{-.5em}
\subsection{Background}
\vspace{-.25em}
Estimating an unknown distribution from its samples
is a fundamental statistical problem arising in many
applications such as modeling language, stocks, weather,
traffic patterns, and many more. It has therefore been
studied for over a century, e.g.~\cite{pea95}.

Consider an unknown univariate distribution $f$ over $ \reals$,
generating $n$ samples $X^n\ed X_1\upto X_n$.
An \emph{estimator} for $f$ is a mapping 
$ \hatf: \reals^n\rightarrow  \reals$.
As in many of the prior works, we evaluate $\hatf$ using the 
$\ell_1$ distance, $\lone{\hatf-f}$.
The $\ell_1$ distance professes several desirable 
properties, including scale and location invariance, 
and provides provable guarantees on the values
of Lipschitz functionals of $ f $~\cite{dev}.

Ideally, we would prefer an
estimator that learns any distribution.
However, arbitrary distributions 
cannot be learned with any number of samples.
Let $u$ be the continuous uniform distribution 
over $[0,1]$. For any number $n$ of samples, 
uniformly select $n^3$ points from
$[0,1]$ and let $p$ be the discrete uniform
distribution over these $n^3$ points. 
Since with high probability collisions do not
occur within samples under either distribution, 
$ u $ and $ p $ cannot be distinguished from 
the uniformly occurring samples. 
As $\lone{u-p}=2$, it follows that for any
estimator $ \hatf $, 
$ \max_{f\in \{u, p\}}\EE\lone{\hatf-f} \gtrsim 1 $.

A common modification, motivated 
by PAC agnostic learning,
assumes that $f$ is close to a natural distribution class $\cC$,
\ignore{Since such as the class of all polynomial distributions over an interval, }
and tries to find the distribution in $\cC$ closest to $f$. 
The following notion of $ \mathrm{OPT}_{\cC}(f) $
considers this lowest distance, and the usual 
\emph{minimax learning rate} of $ \cC$, $ \cR_n(\cC) $, is 
the lowest worst-case expected distance
achieved by any estimator,
\vspace{-.25em}
\[
\mathrm{OPT}_{\cC}(f)
\ed
\inf_{g\in \cC}\lone{f-g}, \ \
\cR_n(\cC)
\ed
\min_{\hatf}
\max_{f\in \cC}\E_{X^{n}\sim f} \lone{\hatf - f}.
\vspace{-.5em}
\]
{As has been considered in~\cite{bousquet19}, 
$\hatf$ is said to be a factor-$c$ approximation for $\cC$ if
\vspace{-.25em}
\[
\EE\lone{\hatf-f}\le c\cdot \OPT_{\cC}(f) + \epsilon_n
\]
where as $n\nearrow \infty$, the \emph{statistical rate},
$\epsilon_n\searrow 0$ at a rate independent of $f$, 
namely, the estimator's error is essentially at most $c$ times the optimal.
Since for $f\in \cC$ has $\OPT_{\cC}(f)=0$, we see that 
$\epsilon_n\ge \cR_n(\cC)$ for any estimator.
}

The key challenge is to obtain such an estimate
for dense approximation classes $ \cC$. 
One such class is the set of degree-$ d $ 
polynomials, $ \cP_d $ and its $ t$-piecewise 
extension, $ \cP_{t,d} $. It is known 
that by tuning the parameters $ t,d $, the bias and variance 
under $ \cP_{t,d}$ can be suitably tailored 
to achieve several in-class minimax rates.
For example, if $ f $ is a 
log-concave distribution, choosing $ t = n^{1/5} $ and $ d=1 $,
$ \OPT_{\cP_{t,d}}(f)+\cR_n(\cP_{t,d}) = \cO(1/n^{2/5})$~\cite{chan14}, 
matching the minimax rate of learning 
log-concave distributions. 
Similarly, minimax rates may be 
attained for many other structured classes including
uni-modal, Gaussian, and mixtures of all three.

The VC dimension, $ \mathrm{VC}(\cC)$, measures
the complexity of a class $ \cC $. 
For many dense classes, including 
$ \cP_{t,d} $, $ \cR_{n}(\cC) = \Theta(\sqrt{\mathrm{VC}(\cC)/n}) $.
For such classes, 
a cross-validation based estimator $\hatf$, such as the minimum
distance based selection~\cite{dev}, across a sufficiently
fine cover of $ \cC $, achieves a factor-3 
approximation to $ \cC$, 
\vspace{-.5em}
\[
\EE \lone{\hatf-f}\le 3\mathrm{OPT}_{\cC}(f)
+\cO(\sqrt{\mathrm{VC}(\cC)/n}).
\]
However, in general, such methods 
might have time complexity exponential in $ n $.
This is especially significant in modern
applications that process a large number of 
samples. \cite{jay17} provided a near-linear
$ \cO(n \log ^{3} n ) $ time
algorithm, $ \ADLS $, that still achieves the same 
factor-3 approximation for $ \cP_{t,d} $ and the 
statistical rate $ \epsilon_n = \cO(\sqrt{t(d+1)/n})$.
However it leaves some important questions unanswered.
\begin{itemize}
\vspace{-.5em}
\item \textbf{Q1:} 
$\ADLS$ shares the same factor-3 approximation as the generic minimum distance selection.
However, for the constant-polynomial class $ \cP_0 $, it is easy to see that the empirical histogram
$ \hatf $ achieves a factor-2 approximation, matching a known lower bound~\cite{dev}.
This raises the question if the factor-3 upper bound can be reduced for higher-degree polynomials as well, and if it can be achieved
with {statistical rate} near the optimal $\sqrt{d(t+1)/n}$. 
\item \textbf{Q2:}  
$ \ADLS $ requires prior knowledge
of the number $ t $ of polynomial pieces, which may be impractical
in real applications.
Even for structured distribution {families}, the $ t $ 
achieving {their minimax rate} can vary significantly. 
For example, for log-concave distributions, 
$ t = \Theta(n^{1/5})$, and for unimodal distributions,
$ t = \Theta(n^{1/3})$.
This raises the question of whether there are estimators that are optimal for $\cP_{t,d}$ simultaneously over all $\forall t\ge 0 $.
\vspace{-.5em}
\end{itemize}
A partial answer for Q1 was provided in \cite{bousquet19} who recently showed that any \emph{finite} class $\cC$ can be approximated with the optimal approximation factor of 2, and with statistical rate $\epsilon_n=\tilde{\cO}\left(|\cC|^{1/5}/n^{2/5}\right)$. 
While this result can be adapted to infinite classes 
	like $\cP_{t,d}$ by
constructing finite covers, as Lemma~\ref{lem:boss} in Appendix~\ref{appen:extra} shows, even for the basic 
single piece quadratic polynomial class $\cP_{2}$, 
this yields $\epsilon_n = \tilde{\cO}(n^{-1/4})\gg
 \Theta(n^{-1/2})=\cR_n({\cP_{2}})$.
And as with the minimum distance selection discussed above, 
the result is only information-theoretic 
without a matching algorithm. 

Q2 can be partially addressed by using cross-validation techniques, for example based on the minimum distance selection that compare results for different 
$t$'s and finds the best. However, {as shown in~\cite{dev}}, this would 
add an extra approximation factor of at least 3, and perhaps even 5 as 
$\ADLS$'s estimates are un-normalized, resulting in $ c = 5\cdot 3 = 15$.
Furthermore this step raises the {statistical rate} 
by an additive $\cO(\log n /\sqrt{n} )$.

$\SURF$ answers both questions in the affirmative. 
Theorem~\ref{thm:polyblack} achieves factors $\le 3$ for all degrees 
$\le 8$ with optimal $\epsilon_n=\cO(\cR_n(\cP_{d}))$.
Corollary~\ref{cor:interhistopt} achieves the same factors and a 
near-optimal $\epsilon_n=\tilde{\cO}(\cR_n(\cP_{t,d}))$ 
for any $t\ge 0$, even unknown, and runs in time $ \cO(n\log^{2}n) $.

The rest of the 
paper is organized as follows. In 
Section~\ref{sec:interval} we describe the
construction of intervals and partitions 
based on statistically equivalent blocks. 
In Section~\ref{sec:black} we present $ \EQ $, 
a polynomial approximation method for any 
queried interval based on a novel empirical 
mass interpolation. 
In Section~\ref{sec:stitch} we explain the
$ \STITCH $ and $ \MERGE $
routines, that respectively combine and compare 
between piecewise polynomial approximations.
We conclude in Section~\ref{sec:experi} with
a detailed comparison of $ \SURF $ and 
$ \ADLS $, and show experimental results that confirm 
the theory and show that $ \SURF $ 
performs well for a variety of distributions.
Proofs of all theorems and lemmas may 
be found in the supplementary material.

%
\vspace{-.5em}
\subsection{Relation to Prior Work}
\vspace{-.5em}
In terms of objectives, $\SURF$ is most closely related to $\ADLS$. 
Briefly, $\SURF$ is simpler, because of which it has a 
$\cO(n\log^2 n)$ time complexity compared to $\cO(n\log^3 n)$,
it is parallizable to 
run in sub-linear time given sorted samples unlike $\ADLS$ that 
uses VC dimension based approaches. 
As mentioned above, it is also more adaptive. 
On the other hand, when $t$ is known in advance, $\ADLS$ achieves a factor-$3$ approximation with optimal $\epsilon_n$.
For a more detailed comparison, see Section~\ref{sec:experi}.

Among the many other methods that have been employed
in distribution estimation, see~\cite{sco12,dev13}, 
$ \SURF $ is inspired by the concept of statistically 
equivalent blocks introduced in~\cite{tukey47,tukey48}.
Distribution estimation methods using this concept partition 
the domain into regions identified by a fixed number
of samples, and perform local estimation on these regions.
These methods have the advantage that they are simple to describe, 
almost always of polynomial time complexity 
in $ n $, and easy to interpret.

The first estimator that used this technique is
found in~\cite{partha61}. 
Expanding on several subsequent works, 
the notable work~\cite{lug96} shows
consistency of a family of equivalent block based
estimators for multivariate distributions.
See~\cite{dev13} for a more extensive treatment of this subject. 
Ours is the first work that 
provides agnostic error guarantees for an 
equivalent block based estimator.

Other popular estimation methods are 
the Kernel, nearest neighbor, MLE, and 
wavelets, see~\cite{sil18}.
Another related method uses splines, for example~\cite{weg83,gu93}. 
While MLE and splines may be used for polynomial estimation,
MLE is intractable in general, and neither
provide agnostic error guarantees.
\vspace{-.5em}
\subsection{Main Results}
\vspace{-.5em}
$ \SURF$ first uses an interpolation routine $ \EQ $ 
that outputs an estimate, $ \hatf_{I,\EQ} \in \cP_d$
for any queried interval $ I $. {Notice that a 
degree-$ d $ polynomial is determined by the measure
it assigns to any $ d+1 $ distinct sub-intervals of $I$.
While $\ADLS$ considers fitting the polynomial that 
minimizes difference in measure to the empirical mass on
the worst set of $d+1$ sub-intervals, we show that for
low-degree polynomials, it suffices to consider 
certain special sub-intervals. Provided in
Lemma~\ref{lem:optnodes}, they 
are functions of $ d $ and are sample independent.
For $d\le 8$, the resulting
estimate is a factor $<3$ approximation to $ \cP_{d} $,
with $ \epsilon_n = \cO(\cR_n(\cP_{d})) $, the optimal 
statistical rate for $ \cP_{d} $.
}
\begin{Theorem}
\label{thm:polyblack}
Given samples $ X^{n-1}\sim f $ for some $ n\ge 128 $,
degree $  d$, and an interval $ I $ with $ n_I$ 
samples within $ I $, $ \EQ $ takes $ \cO(d^\tau+n_I) $ time,
and outputs $ \hatf_{I, \EQ}\in \cP_d $
such that
\vspace{-.5em}
\[
\nonumber
\EE \loneint{I}{\hatf_{I, \EQ}-f}
\le (r_d+1)\cdot \inf_{\poly \in \cP_d} \loneint{I}{\poly - f }
+r_d \cdot \sqrt{\frac{2(d+1)\hatp_I}{\pi n}},
\]
where $ \hatp_{I}\ed(n_I+1)/n $,
$ \loneint{I}{.} $ is the $ \ell_1 $ norm
evaluated on $ I $, $\tau < 2.4$ is the matrix 
inversion exponent, $ r_d $ is a fundamental constant
whose values are $ r_0 = 1, r_1 = 1.25, r_2 \approx 1.42 , r_3 \approx
1.55 $, $ r_4\le 1.675 $, $ r_5\le 1.774 $, 
$ r_6\le 1.857 $, $ r_7\le 1.930 $, $ r_8\le 1.999$
for $ 4\le d \le 8 $.
\end{Theorem}
\vspace{-.75em}
A few remarks are in order.
The additive $ {\cO}(\sqrt{\hatp_I/n}) $ here
is related to the standard deviation in 
the measure associated with an interval 
that has $ \hatp_{I} $ fraction of samples. 
For $ d>8 $, $ r_d>3$ and they may be evaluated
using Lemma~\ref{lem:optnodes}.

The main routine of $ \SURF $, $ \STITCH $, then
calls $ \EQ $ to obtain a piecewise estimate for any 
partition of the domain. $ \STITCH $ uses $ \MERGE $
to compare between the different piecewise estimates. 
By imposing a special binary structure on the space
of partitions, we allow for $ \MERGE $ to efficiently 
make this comparison via 
a divide-and-conquer approach.
This allows $ \STITCH $, 
and in turn $ \SURF $, to output 
$\hatf_{\text{SURF}}$ in $ \cO((d^{\tau}+\log n)n\log n) $ time,
where $\tau$ is the matrix inversion exponent.
$\hatf_{\text{SURF}}$ is a factor-$ (r_d+1)$
approximation for $\cP_{t,d}$ $ \forall t\ge 0 $.
\ignore{As a combined effect, for the case of unknown $ t $,
the best known approximation factor of 
any sub-exponential time algorithm is reduced from
$15$, to $ 2.25,2.42,\ldots, 2.999$, 
respectively, for $ d=1,\ldots,8 $. }
The simplicity of $ \SURF$, both the 
polynomial interpolation and divide-and-conquer,  allow us
to derive all constants explicitly unlike in the previous works. 
This result is summarized below
in Theorem~\ref{thm:interhistopt} and Corollary~\ref{cor:interhistopt}. 
\begin{Theorem}
\label{thm:interhistopt}
Given $X^{n-1}\sim f$ for some $ n\ge 128 $ such 
that $ n $ is a power of $ 2 $, and parameters 
$d\le 8$, $ \alpha>2 $, $ \SURF $ takes 
$ \cO((d^{\tau}+\log n)n\log n) $ time,
and outputs $ \hatf_{\text{SURF}}$
such that w.p. $ \ge 1-\delta $,
\ignore{-.25em} 
\begin{align*}
\lone{\hatf_{\text{SURF}}-f}
\le
\min_{\barI\in \Delta_{\reals}(X^{n-1})}\sum_{I \in \barI} 
&\left(
\frac{(r_d+1) \alpha}{\alpha-2} 
\inf_{\poly \in \cP_d} \loneint{I}{\poly - f }\right.\\
&
\left.+\frac{r_d (\alpha\sqrt2+\sqrt{2}-1)}{(\sqrt{2}-1)^2}
\sqrt{\frac{5(d+1) \hatp_I \log \frac{n}{\delta}}{n}}\right),
\end{align*}
where $ \hatp_{I}$ is the fraction of samples in interval $ I $,
$\Delta_{\reals}(X^n)$ is the collection of all 
partitions of $ \reals $ whose intervals
start and end at a sample point, 
$ \loneint{I}{\cdot} $ is the $ \ell_{1} $ distance
evaluated in interval $ I $,
$ \tau < 2.4 $ is the matrix inversion exponent, 
and $ r_d>0$ is the constant in Theorem~\ref{thm:polyblack}.
\end{Theorem}
\begin{Corollary}
\label{cor:interhistopt}
Running $ \SURF$ with $d\le 8$, 
$\alpha>4$, 
\begin{align*}
\EE\lone{\hatf_{\SURF}-f}
&\le \min_{t\ge 0} \left( 
(r_d+1)\left(1+\frac{4}{\alpha}\right)\cdot \mathrm{OPT}_{\cP_{t,d}}(f)+
\tilde{\cO}\left( \alpha \sqrt{\frac{t\cdot (d+1)}{n}}\right)
\right).
\end{align*}
\end{Corollary}
\vspace{-.5em}
%
\vspace{-.75em}
\section{Intervals and Partitions}
\vspace{-.75em}
\label{sec:interval}
For $n\ge1$, let $X^{(n-1)}\ed \ordst{X}{1}\upto \ordst{X}{n-1}$ 
be the increasingly-sorted values of $X^{n-1}$.
For integers $0\le a<b\le n$, these samples
define intervals on the real line $\reals$,
\[
I_{a,b}
=(-\infty,\ordst{X}{b})\ \text{if} \ a=0, \ 
I_{a,b} = [\ordst{X}{a},\ordst{X}{b}) \ \text{if} \ 0<a<b<n, \
I_{a,b} = [\ordst{X}{a},\infty) \ \text{if} \  b=n.
\]
The \emph{interval-} and \emph{empirical-probabilities} are
$\intp_{a,b} \ed \int_{I_{a,b}}dF, \ \text{and }\
\hatp_{a,b}\ed \frac{b-a}{n}.$
For any $ 0\le a<b\le n $, $ I_{a,b} $ forms a 
\emph{statistically equivalent} block~\cite{tukey47}, wherein
$\intp_{a,b}\sim\Beta(b-a,n-(b-a))$ regardless of $ f $, and $ \intp_{a,b} $ concentrates to $ \hatp_{a,b} $.
\begin{Lemma}
\label{lem:singleconc}
For any $ 0\le a<b\le n $, $\epsilon\ge 0$,
\vspace{-.25em}
\[
\Pr[|\intp_{a,b}-\hatp_{a,b}|\ge \epsilon\sqrt{\hatp_{a,b}}]
\le 
e^{-(n-1)\epsilon^2/2}
+e^{{-(n-1)\epsilon^2 \hatp_{a,b}}/
{(2\hatp_{a,b}+2\epsilon\sqrt{\hatp_{a,b}})}}.
\]
\end{Lemma}
\vspace{-.5em}
We extend this concentration from one interval to many. 
For a fixed $ \epsilon >0 $, let 
$ \cQ_{\epsilon} $ be the event that
\[
\forall \ 0\le a<b\le n ,\ 
|\intp_{a,b}-\hatp_{a,b}|\le\epsilon\sqrt{\hatp_{a,b}}.
\vspace{-.25em}
\]
\begin{Lemma}
\label{lem:multconc}
For any $n\ge 128$ and $ \epsilon\ge 0 $,
\[
P[\cQ_{\epsilon}]\ge 1- {n(n+1)}/{2} \cdot
\left(e^{-(n-1)\epsilon^2/2}
+e^{-(n-1)\epsilon^2/
(2+2\epsilon\sqrt{n})}\right).
\]
\end{Lemma}
\vspace{-.5em}
Notice that $ \cQ_{\epsilon} $ refers to a stronger concentration event
that involves $ \sqrt{\hatp_{a,b}} $ $ \forall 0\le a<b\le n $ and 
standard VC dimension based bounds cannot be 
readily applied to obtain Lemma~\ref{lem:multconc}.

\ignore{-.25em}
A collection of countably many disjoint intervals whose 
union is $\reals$ is said to be a partition of $\reals$.
A distribution $ \barhatp $, consisting
of interval empirical probabilities is called an \emph{empirical distribution}, 
or that each probability in $ \barhatp $ is a multiple of $ 1/n $. 
The set of all empirical distributions is denoted by 
$ \Delta_{\mathrm{emp,n}} $.
Since each $ \hatp\in \barhatp \in \Delta_{\mathrm{\mathrm{emp,n}}} \ge 1/n$,
$ \barhatp $ may be split into its finitely many probabilities as
$\barhatp=(\hatp_1\upto\hatp_k)$. 
These probabilities define a partition if we 
consider the first increasingly sorted $\hatp_1n$ samples, 
the next $\hatp_2n$ samples and so on. 
For $1\le i\le k$, let $\hatcump_{i}\ed\sum_{j=1}^{i-1}\hatp_j$ (note that $\hatcump_1=0$).
The empirical distribution defines the following
\emph{interval partition}:
\[
\nonumber
\barIp{\barhatp}
\ed
(I_{\hatcump_1n,(\hatcump_1+\hatp_1)n},
I_{\hatcump_2n,(\hatcump_2+\hatp_2)n}
\upto
\label{eqn:emppart}
I_{\hatcump_kn,(\hatcump_k+\hatp_k)n}).
\vspace{-1em}
\]
%
\vspace{-.75em}
\section{The Interpolation Routine}
\ignore{-.75em}
\label{sec:black}
This section describes $ \EQ $, which outputs 
an estimate $ \hatf_{I, \EQ} \in \cP_d$ for any queried interval $ I $. WLOG let $I = [0,1] $. A collection, $ \barn_d = (n_0\upto n_{d+1})$
such that $ 0=n_0\le n_1\le\cdots\le n_d\le n_{d+1}= 1 $
is said to be a node partition of $ [0,1] $.
Let $ \cN_d $ be the set of node partitions and
for the set of non-zero polynomials, 
$ \cP_{d}\setminus\{0\} $, define 
$r:\cN_d,\cP_d\rightarrow[1,\infty)$ and 
its suprema
\vspace{-.25em}
\begin{equation}
\label{eqn:polyrat}
r(\barn_d,\poly)
\ed
\frac{\int_{0}^1|\poly|}
{\sum_{i=1}^{d+1}|\int_{n_{i-1}}^{n_i}\poly|}, 
\
r_d(\barn_d)=\sup_{\poly\in\cP_d\setminus \{0\}}r(\barn_d,h).
\end{equation}
\vspace{-.25em}
Notice that $r(\barn_d,\poly)\ge 1$ since the absolute
integral $ \ge $ the sum of absolute areas.
For any node partition $\barn_d\in \cN_d$, let 
$ J_{\barn_d,i} \ed  [n_{i-1},n_i]$, 
$ i\in \{1,\cdots,d+1\} $ so that 
$ \bar{J}_{\barn_d} =(J_{\barn_d,1}\upto J_{\barn_d,d+1})$
partitions $ [0,1] $.
Let $ \hatf_{\barn_d}  \in \cP_{d}$ be the unique 
polynomial whose measure on all $ d+1 $ intervals in $ \barJ_{\barn_{d}} $
matches its empirical mass. It is defined as:
\begin{equation}
\label{eqn:areamatch}
\hatf_{\barn_d} \ed h \in \cP_d : 
\forall i \in \{1,\cdots, d+1\} ,
\int_{n_{i-1}}^{n_{i}} h(z)dz = \hatp_{J_{\barn_d,i}},
\end{equation}
where for $ n_J$ samples that lie within an
interval $ J $, $ \hatp_{J}\ed (n_J+1)/n $. 
Computation of $\hatf_{\barn_d} $ involves a 
calculation of $ d+1 $ empirical masses that takes $ \cO(n_J) $ time, and solving a system of $ d+1 $
linear equations that takes $ \cO(d^{\tau})$ time, 
where $ \tau <2.4 $ is the matrix inversion exponent, for a $ \cO(n_{J}+d^{\tau}) $ run time.
The estimate $ \hatf_{\barn_d} $ corresponding 
to any choice of $\barn_d\in \cN_d$ satisfies the following:
\begin{Lemma}
\label{lem:ratioconst}
For interval $ I=[0,1]$ with empirical probability 
$ \hatp_I $, any $ \barn_d\in\cN_d $, 
and $ \epsilon>0 $, the estimate $ \hatf_{\barn_d} $~\eqref{eqn:areamatch} is such that
under event $ \cQ_{\epsilon} $,
\begin{align*}
\lone{\hatf_{\barn_d}-f} 
&\le 
\left(1+r_d(\barn_d)\right)
\inf_{\poly \in \cP_d}\lone{\poly-f} 
+r_d(\barn_d) \epsilon \sqrt{(d+1)\hatp_I}.
\end{align*}
\end{Lemma}
\vspace{-.5em}
In Lemma~\ref{lem:zeroish}, we show that for any 
$ \barn_d\in \cN_{d}$, there exists an 
$ r_d(\barn_d)$ achieving $\poly \in \cP_{d} $, 
and that it belongs to a special set, $ \cP_{\barn_d} \subseteq \cP_d$,
\begin{align*}
\cP_{\barn_d}
&\ed 
\Big\{\poly \in \cP_d : \exists 
i_1\in\{1\upto d+1\} :
\forall i\in \{1 \upto d+1\} \setminus \{i_1\},
\int_{n_{i-1}}^{n_i}\poly = 0\
\Big\}.
\vspace{-.5em}
\end{align*}
In words, $\cP_{\barn_d}$ is the set of
polynomials that has a non-zero area in at most 
one $ I \in \barI_{\barn_d}$.
\ignore{$ \poly\in\cP_d $ to be considered in when 
finding $ \barn_d \in \cN_d $ with a bounded 
$ r(\barn_d) = \sup_{\poly\in\cP_d}r(\barn_d,h)$.
 Formally,
Lemma~\ref{lem:zeroish} shows that the 
largest ratio over $ \poly \in \cP_d $ is attained 
by some $ \poly \in \cP_{\barn_d}$.}
\begin{Lemma}
\label{lem:zeroish}
For any degree-$d$ and $ \barn_d\in\cN_d $,
\[r_d(\barn_d)=\sup_{\poly\in\cP_d}r(\barn_d,h)
=\max_{\poly\in\cP_{\barn_d}}r(\barn_d,h).\]
\end{Lemma}
\vspace{-.75em}
Let the smallest $ r_d(\barn_d) $ be denoted by
$
r^{\star}_d \ed \inf_{\barn_d\in \cN_d} r_d(\barn_d)$.
Lemma~\ref{lem:optnodes} shows that there exists
an $ \barn_d $ that attains the infimum. It is denoted by
$ \barn_d^{\star} 
= \text{arg}\min_{\barn_d\in \cN_d} r_d(\barn_d).$
For $ d\le 3 $, we calculate $ r^{\star}_d $ and $ \barn^{\star}_d $. 
For $ 4\le d\le 8 $ we find a $ \barn_d \in \cN_d$ 
such that the corresponding $ r_d(\barn_d) < 2$.
\begin{Lemma}
\label{lem:optnodes}
For $d\le 3 $, there exists a node collection 
$ \barn^{\star}_d $
that achieves $ r^{\star}_d $. These,
and their respective $ r^{\star}_d $ are given by
\vspace{-.5em}
\begin{center}
\begin{tabular}{ |c|c|c| } 
\hline
$ d $ & $ \barn^{\star}_d $  & $r^{\star}_d$ \\ 
\hline
0 & $ (0,1) $ & $ 1 $ \\
\hline
1 & $ (0,0.5,1) $ & $ 1.25 $ \\
\hline
2 & $ \approx (0, 0.2599, 0.7401, 1) $ & $ \approx 1.42 $ \\ 
\hline
3 & $ \approx(0, 0.1548, 0.5, 0.8452, 1) $& $\approx 1.56 $\\ 
\hline
\end{tabular}
\end{center}
Denoting $\barn^{\star}_2 = (0,\alpha_0,1-\alpha_0, 1)  $, 
and $ \barn^{\star}_3 = (0, \beta_0, 0.5, 1-\beta_0, 1) $,
the exact values of $ \alpha _0 $, $ \beta_0 $, 
are obtained as roots to a degree-$14$ and 
degree-$ 69 $ polynomial that we explicitly provide.
For degrees $ 4\le d\le 8 $, the following 
$ \barn_d\in \cN_d$ and $ r_d(\barn_d) $ 
provide upper bounds on $ r_d^{\star} $.
\begin{center}
\begin{tabular}{ |c|c|c| } 
\hline
$ d $ & $ \barn_d $  & $r_d(\barn_d)$ \\ 
\hline
4 & $ (0, 0.1015, 0.348, 0.652, 0.8985,1)$ & 
$ < 1.675$ \\
\hline
5 & $ ( 0 ,   0.071 ,   0.254 ,   0.5  ,   0.746,   0.929  ,  1) $ & $ <  1.774 $ \\ 
\hline
6 & $ (0,    0.053,    0.192,    0.390,    0.610,   0.808,   0.947,    1) $& $ < 1.857 $\\ 
\hline
7 & $ (0, 0.0405   , 0.149 ,  0.310,   0.5,  0.690,    0.851, 0.9595,    1) $ & $ < 1.930 $ \\
\hline
8 & $  (0,    0.032,   0.119,    0.252,    0.414,   0.586,   0.749,    0.881,   0.968,   1) $ & $ < 1.999 $ \\
\hline
\end{tabular}
\end{center}
\end{Lemma}
\vspace{-.75em}
For a given interval $ I $ and $ d\le 8 $,
$ \EQ $ first scales and shifts $ I $ to obtain
$ [0,1] $. It then constructs $\hatf_{\barn_d}$ using
the $ \barn_d$ in Lemma~\ref{lem:optnodes}.
The output $\hatf_{I, \EQ}$ is the 
re-scaled-shifted $\hatf_{\barn_d}$.
\vspace{-.75em}
\section{The Compare and Merge Routines}
\ignore{Change the title!!!}
\vspace{-.75em}
\label{sec:stitch}
This section presents $ \STITCH $ and $ \MERGE $, the main routines 
of $ \SURF $. For any contiguous collection of intervals $\barI$, let $ \hatf_{\barI,\EQ} $ 
be the piecewise polynomial estimate consisting of $ \hatf_{I, \EQ} \in \cP_d$
given by $ \EQ $ in each $ I\in \barI $. The key idea in 
$ \SURF $ is to separate interval partitions into a binary hierarchy, effectively 
allowing a comparison of all the
superpolynomially many (in $ n $) 
estimates corresponding to the different interval partitions, 
but by using only $ \tilde{\cO}(n) $ comparisons. 

Recall that $ n $ here a power of $ 2 $ and define the integer
$D \ed \log_{2} n.$
An empirical distribution, $ \barhatp \in \Delta_{\mathrm{emp,n}} $,
is called a \emph{binary} distribution if each of its probability values 
take the form $ 1/2^d $, for some integer 
$0\le d\le D$. The corresponding interval partition, 
$ \barI_{\barhatp} $, is said to be a \emph{binary partition}.
\begin{align*}
\Delta_{\mathrm{bin,n}}
&\ed 
\{\barhatp\in \Delta_{\mathrm{emp,n}}
:\forall \hatp\in\barhatp, \hatp=1/2^{\nu(\hatp)}, 0\le \nu(q)\le D, \ 
\nu(q)\in \mathbb{Z}\}.
\end{align*}
For example $ \barhatp=(1)$, $\barhatp=(1/2,1/4,1/4)$,  
$\barhatp=(1/4,1/8,1/8,1/2) $ are binary
distributions. Similarly, $\baru{n} = \baru{2^{\log n}}$
is also a binary distribution since $ n $ here is 
a power of 2 (assume $ n\ge 8 $ so that they 
are all in $\Delta_{\textrm{emp,n}}$). 
Lemma~\ref{lem:binarycomp} shows that $\Delta_{\mathrm{bin,n}}$ retains
most of the approximating power of $\Delta_{\mathrm{emp,n}}$. In particular, that
for any $ \barhatp \in \Delta_{\mathrm{emp,n}}$, there exists a binary distribution $\barhatp' \in \Delta_{\mathrm{bin,n}}$ such that
$ \barI_{\barhatp'} $ has
a smaller bias than $ \barI_{\barhatp} $, while its
deviation under the 
concentration event, $ \cQ_{\epsilon} $,
is larger by less than a factor of $ 1/(\sqrt{2}-1) $.
\begin{Lemma}
\label{lem:binarycomp}
For any empirical distribution
$ \barhatp \in \Delta_{\mathrm{emp,n}}$, 
there exists $  \barhatp' \in \Delta_{\mathrm{bin,n}}$ 
such that
\vspace{-.5em}
\[
\lone{\starf_{\barI_{\barhatp'}}-f} \le
\lone{\starf_{\barI_{\barhatp}}-f}, \
\sum_{\hatp \in \barhatp'}
\epsilon\sqrt{\hatp}
\le
\sum_{\hatp\in \barhatp}
\frac{1}{\sqrt{2}-1}\epsilon\sqrt{\hatp},
\vspace{-.25em}
\]
where for any $d>0$,
$ \starf_{\barI} $ is the piecewise degree-$d$
polynomial closest to $ f $ on the partition $ \barI $.
\end{Lemma}
\vspace{-.5em}
For a fixed $ \barp\in \Delta_{\mathrm{bin,n}}$,
let $\Delta_{\mathrm{bin,n}, \le \barp}$ be the 
set of binary distributions such that for any 
$\barhatp \in \Delta_{\mathrm{bin,n}, \le \barp}$,
each $ I_1\in \barI_{\barhatp} $ is contained 
in some $I_2 \in \barI_{\barp} $.
\vspace{-.4em}
\begin{equation}
\label{eqn:nestbin}
\Delta_{\mathrm{bin,n}, \le \barp } \ed 
\{\barhatp\in \Delta_{\mathrm{bin,n}}
:\forall I_1 \in \barI_{\barhatp}, \
\exists I_2 \in \barI_{\barp}, \
I_1\subseteq I_2
\}.
\end{equation}
For example if $ \barp=(1/2,1/4,1/4) $ is the 
binary distribution, $(1/4,1/4,1/8, 1/8,1/4)$, 
$(1/2,1/4,1/8,1/8) \in \Delta_{\mathrm{bin,n},\le \barp}$, whereas
$ (1/2,1/2) \notin  \Delta_{\mathrm{bin,n},\le \barp} $.
\vspace{-.5em}
\subsection{The $\STITCH$ Routine}
\vspace{-.5em}
The $ \STITCH $ routine  
operates in $i\in\{1\upto D\} $ steps (recall $D=\log_2n$)
where at the end of each step $ i $, $ \STITCH $ holds onto 
a binary distribution $ \hatp_{i} $. At at the last 
step $ i=D $, $ \SURF $ outputs the piecewise 
estimate on the partition given by $\barhatp_{D} $, i.e. 
$ \hatf_{\text{SURF}}  
=\hatf_{\barI_{\barhatp_{D}},\EQ } $. Let
\vspace{-.25em}
\[  D(i)\ed D-i \text{ and let }
\bar{u}_i \ed \left({1}/{2^{D(i)}}\upto {1}/{2^{D(i)}}\right).
\]
Initialize $\barhatp_0\gets \baru{n}$. Start with $ i =1  $ and 
assign $ \barhats \gets \barhatp_{i-1}$.
Throughout its run $\STITCH$ maintains
$\barhats = \barhatp_{i-1}\in \Delta_{\mathrm{bin,n},\le 
\bar{u}_i}$. For instance this holds for $i=1$ since 
$ \bar{u}_1 = (2/n\upto 2/n)$. 
$\STITCH$ considers merging the probability values
in $\barhats$ to match it with $\bar{u}_i$.
For example if at step $i=D-1$, $\barhats= (1/8,1/8,1/4,1/4,1/4)$,
it considers merging $(1/8,1/8,1/4)$
and $(1/4,1/4)$ to obtain $\bar{u}_{D-1}=(1/2,1/2)$.

This decision is made by invoking the $\MERGE$ routine 
on intervals corresponding to the merged 
probability value. 
In this case $\MERGE$ is called
on intervals $\barI \in \barI_{\barhats}$ corresponding to 
$(1/8,1/8,1/4)$ and $(1/4,1/4)$ respectively, along with
the tuning parameter $\gamma$,
\[\gamma \ed \alpha\cdot r_d \cdot \epsilon \sqrt{d+1}.\]
While $\MERGE$ decides to merge depending on the increment
in bias on the merged interval versus the decrease in variance,
$\gamma$ tunes this trade-off. 
A large $ \gamma  $ results in a decision to merge while
a small $ \gamma  $ has the opposite effect.
If $\MERGE(\barI,\gamma)\le 0$ the probabilities
in $\barhats$ corresponding to $\barI$ are merged and copied 
into $\barhatp_i$. Otherwise they are copied as is into $\barhatp_i$. See Appendix~\ref{appen:alg1} for a detailed description.

At each step $ i\in \{1\upto D\} $, $ \STITCH $ calls 
$ \MERGE $ on $2^{D(i)} $ intervals,
each consisting of $ 2^i $ samples. 
Thus each step of $ \STITCH $ takes
$ \cO(2^{D(i)}\cdot (d^{\tau}+\log (2^i)) \cdot 
2^i) = \cO((d^{\tau}+\log n )2^D)$ time. 
The total time complexity is therefore 
$\cO((d^{\tau}+\log n )2^D D) = \cO((d^{\tau}+\log n)n\log n ) $.

\vspace{-.5em}
\subsection{The $\MERGE$ Routine}
\vspace{-.5em}
$ \MERGE $ receives an interval partition $\barI$ 
consisting of $m$
samples and the parameter $\gamma$ as input,
and returns a real value that indicates its decision 
to merge the probabilities under $\barI$. Let $\barhats$
be the set of empirical probabilities corresponding to 
$\barI$. Let the merged interval 
be $I$ and let $\hatf = \hatf_{\EQ,I}$ be 
the polynomial estimate
on $I$.

For simplicity suppose $\barI = (I_1,I_2)$ 
with 
empirical mass $ \hats_{I_{1}} $, $ \hats_{I_{2}} $ respectively,
and let $ \OPT_{I,\cP_{t,d}}(f) =\min_{h\in \cP_{t,d}}\loneint{I}{\poly - f}$. 
If $\barI$ is merged, observe that the bias
$ \OPT_{I,\cP_{t,d}}(f)\ge \OPT_{I_{1},\cP_{t,d}}(f)+\OPT_{I_{2},\cP_{t,d}}(f) $ 
increases but since
$  \hats_{I}=\hats_{I_1}+\hats_{I_2} $,
$
\sqrt{\hats_{I}}
\le 
\sqrt{\hats_{I_{1}}}+
\sqrt{\hats_{I_{2}}},
$
resulting in a smaller $ \epsilon $-deviation under event $\cQ_\epsilon$ in Lemma~\ref{lem:multconc}.
Consider their difference parameterized by the constant 
$ \gamma $,
\[  \mu'_{\gamma}(f)  \ed (\OPT_{I_{1},\cP_{t,d}}(f)+\OPT_{I_{2},\cP_{t,d}}(f) -\OPT_{I,\cP_{t,d}}(f) )
-\gamma(\sqrt{\hats_{I_{1}}}+
\sqrt{\hats_{I_{2}}}-\sqrt{\hats_{I}}).
\]
If $ \mu'_{\gamma}(f)  \le 0 $, it indicates that 
the overall $\ell_1$ error is smaller 
under the merged $ I $. While
$  \mu'_{\gamma}(f) $ cannot
be evaluated without access to the underlying  $f$, 
we use a proxy, 
$ \mu_{\barI,\gamma}(f)$ that is defined next.

Normalize $\barhats$ so that it is a distribution, and consider 
$ \barp\in \Delta_{\mathrm{bin,m}}$ such that 
$\barhats \in \Delta_{\mathrm{bin,m},\le \barp} $ and 
the piecewise estimate 
on $ \barI_{\barp} $, i.e. $ \hatf_{\barI_{\barp},\EQ} $. Define
$
\del_{\barI_{\barp}}(\hatf)
\ed
\loneint{\barI_{\barp}}{\hatf_{\barI_{\barp},\EQ}
	-\hatf},\
\thr_{\barp, \gamma}\ed 
\sum_{p\in \barp}
\gamma \sqrt{p},
$
\vspace{-.3em}
\[
\lemerge_{\barI_{\barhats},\gamma}(\hatf)
\ed
\max_{\barp: \barhats \in \Delta_{\mathrm{bin,m}, \le \barp }}
\del_{\barI_{\barp}}(\hatf)-\thr_{\barp, \gamma}.
\vspace{-.2em}
\]
$ \MERGE $ returns $ \lemerge_{\barI_{\barhats},\gamma}(\hatf) $
via a divide-and-conquer based implementation, and results in $  \cO((d^{\tau}+\log m)m) $ time. A detailed description is provided in Appendix~\ref{appen:algo}.
%
%
Lemma~\ref{lem:master} shows that under 
event $ \cQ_{\epsilon} $, 
$ \hatf_{\SURF }$ is within a constant 
factor of the best piecewise polynomial approximation 
over any binary partition, plus its deviation in probability 
under $ \cQ_{\epsilon} $ times $ \cO(\sqrt{d+1}) $.
\begin{Lemma}
\label{lem:master}
Given samples $ X^{n-1}\sim f $, for some $ n $ that 
is a power of $ 2 $, degree $ d\le 8 $ and
the threshold $ \alpha > 2$, $ \SURF$ 
outputs $ \hatf_{\text{SURF}} $ in time 
$ \cO((d^\tau+\log n)n\log n) $ such that under event 
$ \cQ_{\epsilon} $,
\begin{align*}
\lone{\hatf_{\text{SURF}}-f}
\le
\min_{\barp \in \Delta_{\mathrm{bin,n}}(X^{n-1})}
\sum_{I \in \barI_{\barp}} 
&\left(
\frac{(r_d+1) \alpha}{\alpha-2} 
\inf_{\poly \in \cP_d}\loneint{I}{\poly-f}\right.\\ &
\left.+\frac{r_d (\alpha\sqrt2+\sqrt{2}-1)}{\sqrt{2}-1}
\epsilon \sqrt{(d+1)\hatp_{I}}\right),
\vspace{-.5em}
\end{align*}
where $ q_I $ is the empirical mass under interval $ I $,
$ r_d $ is the constant in 
Theorem~\ref{thm:polyblack}.
\end{Lemma}
\vspace{-.75em}
 \vspace{-.75em}
\section{Comparison and Experiments}
\label{sec:experi}
\vspace{-.75em}
We compare the factor improvement of 
$\SURF$ with $\ADLS$, expand on larger degrees-$d$ polynomial approximation, and in particular, address learning Gaussians optimally. 
We also describe how $\SURF$ benefits from its local nature, enabling a distributed computation.
Our experiments show that $\SURF$ is more adaptive than $\ADLS$, and perform additional experiments on both synthetic and real datasets.

The following table compares $ \SURF $ with $ \ADLS $
in terms of the expected error.  
For $ d\le 8 $, $ r_d\in [2,3) $ 
is the factor in 
Theorem~\ref{thm:polyblack}, 
and $ \tau,\omega\in [2, 2.4] $ are constants. 
We achieve a lesser factor approximation 
at nearly the optimal statistical rate, 
with an improved time complexity in both $n$ and $d$.
\begin{center}
\begin{tabular}{ |c|c|c| } 
\hline \xrowht{7pt}
& $ \SURF$  
& $ \ADLS $ \\ 
\hline \xrowht{9pt}
$ \cP_{d}$ 
& $r_d \, \OPT_{\cP_{t,d}}(f) +\sqrt{\frac{2d}{\pi n}}$
& $3 \OPT_{\cP_{t,d}}(f) +\cO\!\left(\!\sqrt{\frac{d}{n}}\right)$  \\
\hline \xrowht{9pt}
$ \cP_{t,d}$ and known $ t $ 
& $r_d \, \OPT_{\cP_{t,d}}(f) +\cO\!\left(\!\sqrt{\frac{t(d+1)\log n}{n}}\right)$
& $3   \OPT_{\cP_{t,d}}(f) 
+\cO\!\left(\!\sqrt{\frac{t(d+1)}{n}}\right)$  \\
\hline \xrowht{9pt}
$ \cP_{t,d}$ and unknown $ t $ 
& $r_d \min_{t\ge 0} \left(\OPT_{\cP_{t,d}}(f) 
\right.
$
& $15 \min_{t\ge 0} \left(\OPT_{\cP_{t,d}}(f) 
\right.
$  \\
\xrowht{10pt}
& $+\left.\cO\left(\sqrt{\frac{t(d+1)\log n}{n}}\right)\right)$
& $+\left.\cO\left(\sqrt{\frac{t(d+1)}{n}}\right)\right)+ \cO\left(\frac{\log n}{\sqrt{n}}\right)$   \\
\hline \xrowht{8pt}
Time complexity 
& $ \cO(n\log^{2}nd^{\tau}) $
&  $ \cO(n\log^3nd^{3+\omega}) $\\ 
\hline 
\end{tabular}
\end{center}
While for $ d>8 $, $ \SURF $ does not improve the approximation factor below $<3$, we note that polynomial approximations of larger 
degrees exhibit oscillatory behavior, for example around the edges 
when approximating a pulse. 
Called the Runge 
phenomenon~\cite{tref13}, this may result in 
an  unbounded $\ell_p$ distance for $ p>1 $.
In this scenario it may be preferred to use a 
lower degree polynomial, but with an appropriately 
large $ t $. 
Consider the important case when $ f $ is 
a Gaussian distribution. As shown in Lemma~\ref{lem:gauss}, 
$\OPT_{\cP_{t,d}}(f)=\cO(1/t^{d-1})$.
Using the fact that $ \epsilon_n
= \tilde{\cO}(\sqrt{t(d+1)/n})$
and minimizing $\OPT_{\cP_{t,d}}(f)+\epsilon_n$ over
$t$ for a fixed $ d $,
we obtain $ \lone{\hatf_{\SURF}-f}=
\tilde{\cO}((d+1)/n)^{\frac12-\frac1{4d-2}}$. 
Even for an astronomical 
$ n = 2^{100} $ samples, choosing $ d = 8  $ 
ensures that $ n^{\frac1{4d-2}}\le 11  $. Thus 
in almost all scenarios of practical interest
we nearly match (upto a $\sqrt{\log n}$ factor) 
the minimax rate 
$ \cO(1/n)^{\frac12} $ of learning Gaussians.
While $ \ADLS $ avoids this factor of 
$n^{\frac1{4d-2}}$, they do so by using 
$ d=\cO(\log n)$ which may present the above drawbacks. For degrees that are even 
larger, the $ \Omega(d^{5}n\log^3 n) $
time taken by $ \ADLS $ may make it impractical.

In terms of time complexity, $ \SURF $ benefits from its local 
nature, enabling a distributed computation. As detailed in Appendix~\ref{appen:algomain}, if provided with
pre-sorted samples, a known $ t $ and memory $ m \ge t$, 
it can be adapted to run in 
time $\cO((d^{\tau}+\log n)n\max\{1/t,\log n/m\}) \ll \cO(n)$, 
if $t\approx n$.
\begin{figure}
	\centering
	\subfigure[$.4\text{Beta}(.8, 4)+$$.6\text{Beta}(2, 2) $]{\begin{tikzpicture}[scale=.55]
\begin{axis}[
    xmin=0, xmax=66000,
    ymin=0, ymax=0.45,
    xtick={ 0, 10000,20000, 30000,40000,50000,60000},
    yticklabels={0,0,0.10,0.2,0.3,0.4,.5},
    legend pos=north east,
    ymajorgrids=true,
    grid style=dashed,
]
\addplot[
color=black,
mark=oplus*,
style=densely dotted
]
coordinates {
	(1000,0.41129029)(2000,0.31566993)(5000,0.21024782)
	(10000,0.15176184)(20000,0.10975702)(50000,0.06909804)(100000,0.04932885)
};
\addlegendentry{$ \mathrm{ADLS}, t=60 $}
\addplot[
    color=purple,
    mark=oplus*,
	style=densely dotted
    ]
    coordinates {
	(1000,0.35273703)(2000,0.2658395)(5000,0.17467638)
	(10000,0.12670596)(20000,0.08974599)(50000,0.05735348)(100000,0.04003552)
    };
    \addlegendentry{$ \mathrm{ADLS}, t=40 $}
\addplot[
    color=olive,
    mark=triangle*,
	style=solid
    ]
    coordinates {
(1000,.26833729)(2000,0.19596455)(5000,0.12682624)
	(10000,0.08992519)(20000,0.06356332)(50000,0.04108105)(100000,0.02863543)
    };
    \addlegendentry{$ \mathrm{ADLS}, t=20 $}
\addplot[
    color=red,
    mark=square*,
	style=densely dotted
    ]
    coordinates {
	(1000,0.19988313)(2000,0.14288181)(5000,0.09270712)
	(10000,0.06511393)(20000,0.04701249)(50000,0.02991442)(100000,0.02137126)
    };
    \addlegendentry{$ \mathrm{ADLS}, t=10 $}
\addplot[
    color=black,
    mark=triangle*,
	style=densely dashed
    ]
    coordinates {
	(1000,0.14325111)(2000,0.10507772)(5000,0.07020512)
	(10000,0.0515054)(20000,0.036336)(50000,0.02416713)(100000,0.01796713)
    };
    \addlegendentry{$ \mathrm{ADLS}, t=5 $}
\addplot[
    color=blue,
    mark=oplus*,
	style=solid
    ]
    coordinates {
    (512,0.1113)(1024,0.0959)(2048,0.0863)
	(4096,0.0689)(8192,0.0416)(16384,0.0327)
	(32768,0.0225)(65536,0.0165)
    };
    \addlegendentry{$ \mathrm{SURF} $}
\end{axis}
\end{tikzpicture}}
	\hfill
	\subfigure[$.7\text{Gam}(2, 2)+$$.3\text{Gam}(7.5, 1) $ ]{\begin{tikzpicture}[scale=.55]
\begin{axis}[
xmin=0, xmax=66000,
ymin=0, ymax=0.45,
xtick={ 0, 10000,20000, 30000,40000,50000,60000},
yticklabels={0,0,0.10,0.2,0.3,0.4,.5},
legend pos=north east,
ymajorgrids=true,
grid style=dashed,
]
\addplot[
color=black,
mark=oplus*,
style=densely dotted
]
coordinates {
	(1000,0.40957123)(2000,0.31347938)(5000,0.21093403)
	(10000,0.15181531)(20000,0.1067836)(50000,0.0689115)(100000,0.04892543)
};
\addlegendentry{$ \mathrm{ADLS}, t=60 $}
\addplot[
    color=purple,
    mark=oplus*,
	style=densely dotted
    ]
    coordinates {
	(1000,0.35356818)(2000,0.26368842)(5000,0.17421362)
	(10000,0.12519919)(20000,0.08865087)(50000,0.05664164)(100000,0.04027485)
    };
    \addlegendentry{$ \mathrm{ADLS}, t=40 $}
\addplot[
    color=olive,
    mark=triangle*,
	style=solid
    ]
    coordinates {
(1000,.26646272)(2000,0.19445114)(5000,0.12502172)
	(10000,0.08849658)(20000,0.06368279)(50000,0.03998866)(100000,0.02899073)
    };
    \addlegendentry{$ \mathrm{ADLS}, t=20 $}
\addplot[
    color=red,
    mark=square*,
	style=densely dotted
    ]
    coordinates {
	(1000,0.19215706)(2000,0.14309746)(5000,0.08978203)
	(10000,0.06416699)(20000,0.04583009)(50000,0.02981915)(100000,0.02135969)
	};
\addlegendentry{$ \mathrm{ADLS}, t=10 $}
\addplot[
    color=black,
    mark=triangle*,
	style=densely dashed
    ]
    coordinates {
	(1000,0.14402419)(2000,0.09954654)(5000,0.06823559)
	(10000,0.04724663)(20000,0.03568414)(50000,0.02389949)(100000,0.01766414)
    };
    \addlegendentry{$ \mathrm{ADLS}, t=5 $}
\addplot[
    color=blue,
    mark=oplus*,
	style=solid
    ]
    coordinates {
    (512,0.1526)(1024,0.1186)(2048,0.0874)
	(4096,0.0638)(8192,0.0464)(16384,0.0330)
	(32768,0.0232)(65536,0.0158)
    };
    \addlegendentry{$ \mathrm{SURF} $}
\end{axis}
\end{tikzpicture}}
	\hfill
	\subfigure[.65$\cN$(-.45,$.15^2$)+.35$\cN$(.3,$.2^2$) ]{\begin{tikzpicture}[scale=.55]
\begin{axis}[
xmin=0, xmax=66000,
ymin=0, ymax=0.45,
xtick={ 0, 10000,20000, 30000,40000,50000,60000},
yticklabels={0,0,0.10,0.2,0.3,0.4,.5},
legend pos=north east,
ymajorgrids=true,
grid style=dashed,
]
\addplot[
color=black,
mark=oplus*,
style=densely dotted
]
coordinates {
	(1000,0.41175876)(2000,0.31447587)(5000,0.21200917)
	(10000,0.1518456)(20000,0.10766436)(50000,0.06875379)(100000,0.04877435)
};
\addlegendentry{$ \mathrm{ADLS}, t=60 $}
\addplot[
    color=purple,
    mark=oplus*,
	style=densely dotted
    ]
    coordinates {
	(1000,0.35143194)(2000,0.26255467)(5000,0.17301745)
	(10000,0.12435304)(20000,0.08918697)(50000,0.05658025)(100000,0.03956338)
    };
    \addlegendentry{$ \mathrm{ADLS}, t=40 $}
\addplot[
    color=olive,
    mark=triangle*,
	style=solid
    ]
    coordinates {
(1000,.26669219)(2000,0.19148383)(5000,0.12435307)
	(10000,0.09116633)(20000,0.06413761)(50000,0.04051442)(100000,0.02940496)
    };
    \addlegendentry{$ \mathrm{ADLS}, t=20 $}
\addplot[
    color=red,
    mark=square*,
	style=densely dotted
    ]
    coordinates {
	(1000,0.19713028)(2000,0.14383183)(5000,0.09048367)
	(10000,0.06666971)(20000,0.04703464)(50000,0.03183507)(100000,0.02351853)
};
    \addlegendentry{$ \mathrm{ADLS}, t=10 $}
\addplot[
    color=black,
    mark=triangle*,
	style=densely dashed
    ]
    coordinates {
	(1000,0.1444273)(2000,0.10918461)(5000,0.07411804)
	(10000,0.05527344)(20000,0.0445799)(50000,0.03693705)(100000,0.03249407)
    };
    \addlegendentry{$ \mathrm{ADLS}, t=5 $}
\addplot[
    color=blue,
    mark=oplus*,
	style=solid
    ]
    coordinates {
    (512,.1840)(1024,0.1409)(2048,0.1035)
	(4096,0.0826)(8192,0.0637)(16384,0.0428)
	(32768,0.0347)(65536, 0.0251)
    };
    \addlegendentry{$ \mathrm{SURF} $}
\end{axis}
\end{tikzpicture}}
	\caption{$\ell_{1} $ error versus number of samples of 
		piece-wise linear $ \SURF $ and $ \ADLS $.}
    \vspace{-.5em}
	\label{plot:compare}
\end{figure}
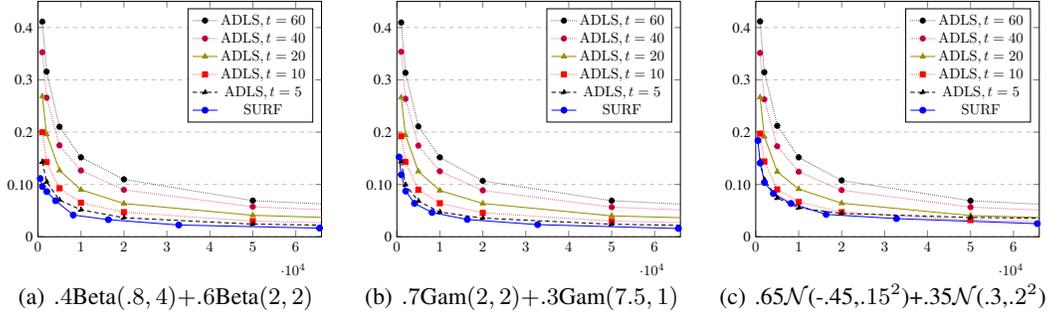
\begin{figure*}
	\centering
	\subfigure[Densities $ f_1 $, $ f_2 $, $ f_3 $.]{\begin{tikzpicture}[scale=.4]
\begin{axis}[
    title={$ \text{Beta} $ Mixture Densities},
    axis lines = left,
    xlabel = $x$,
    ylabel = {},
]

\addplot [
    domain=0:1, 
    samples=100, 
    color=red,
	style=dashed
]
{0.4*60*x^2*(1-x)^3+0.6*30*x^4*(1-x)};
\addlegendentry{$ f_1(x) $}

\addplot [
    domain=0:1, 
    samples=100, 
    color=blue,
	style=solid
    ]
    {0.4*660*x^9*(1-x)^2+0.6*72*x*(1-x)^7};
\addlegendentry{$f_2(x) $}
 
\addplot [
    domain=0:1, 
    samples=100, 
    color=black,
	style={densely dotted},
    ]
    {2722*x^5*(1-x)^5};
\addlegendentry{$ f_3(x)$}
\end{axis}
\end{tikzpicture}\label{aa}}
	\hfill
	\subfigure[$ \ell_1 $ error with $ d=1 $.]{
\begin{tikzpicture}[scale=.4]
\begin{axis}[
    title={Linear $ \SURF $},
    xlabel={Number of Samples },
    xmin=0, xmax=35000,
    ymin=0, ymax=0.17,
    xtick={0,5000,10000,15000,20000,25000,30000,35000},
    yticklabels={0,0,0.05,0.10,0.15},
    legend pos=north east,
    ymajorgrids=true,
    grid style=densely dashed,
]
 
\addplot[
    color=red,
    mark=square*,
	style=dashed
    ]
    coordinates {
    (512,.12)(1024,.095)(2048,0.0789)
	(4096,0.06)(8192,0.048)(16384,0.0352)
	(32768,0.0258)
    };
    \addlegendentry{$ f_1 $} 

\addplot[
    color=blue,
    mark=triangle*,
    ]
    coordinates {
    (512,.16)(1024,.1196)(2048,0.0896)
	(4096,0.0731)(8192,0.0525)(16384,0.0392)
	(32768,0.0303)
    };
    \addlegendentry{$ f_2 $} 
\addplot[
    color=black,
    mark=oplus*,
	style={densely dotted}
    ]
    coordinates {
    (512,.1723)(1024,.1103)(2048,0.0917)
	(4096,0.0644)(8192,0.0461)(16384,0.0346)
	(32768,0.0285)
    };
    \addlegendentry{$ f_3 $} 
\end{axis}
\end{tikzpicture}\label{bb}}
	\hfill
	\subfigure[$ \ell_1 $ error with $ d=2 $.]{
\begin{tikzpicture}[scale=.4]
\begin{axis}[
    title={Quadratic $\SURF $},
    xlabel={Number of Samples },
    xmin=0, xmax=35000,
    ymin=0, ymax=0.17,
    xtick={0,5000,10000,15000,20000,25000,30000,35000},
    yticklabels={0,0,0.05,0.10,0.15},
    legend pos=north east,
    ymajorgrids=true,
    grid style=dashed,
]
 
\addplot[
    color=red,
    mark=square*,
	style=densely dashed
    ]
    coordinates {
    (512,0.125)(1024, 0.07 )(2048,0.054)
	(4096,0.0411)(8192,0.0388)(16384,0.0323)
	(32768,0.0240)
    };
    \addlegendentry{$ f_1 $} 

\addplot[
    color=blue,
    mark=triangle*,
    ]
    coordinates {
    (512,.1191)(1024,0.0950)(2048,0.0778)
	(4096,0.0541)(8192,0.0384)(16384,0.0327)
	(32768,0.0251)
    };
    \addlegendentry{$ f_2 $} 
\addplot[
    color=black,
    mark=oplus*,
	style={densely dotted}
    ]
    coordinates {
    (512,.1222)(1024,.1085)(2048,0.0819)
	(4096,0.0627)(8192,0.0396)(16384,0.0350)
	(32768,0.0288)
    };
    \addlegendentry{$ f_3 $} 
\end{axis}
\end{tikzpicture}\label{cc}}
	\hfill
	\subfigure[$ \ell_1 $ error with $ d=3 $.]{
\begin{tikzpicture}[scale=.4]
\begin{axis}[
    title={Cubic $\SURF $},
    xlabel={Number of Samples },
    xmin=0, xmax=35000,
    ymin=0, ymax=0.17,
    xtick={0,5000,10000,15000,20000,25000,30000,35000},
    yticklabels={0,0,0.05,0.10,0.15},
    legend pos=north east,
    ymajorgrids=true,
    grid style=dashed,
]
 
\addplot[
    color=red,
    mark=square*,
	style=densely dashed
    ]
    coordinates {
    (512,0.0890)(1024, 0.0826 )(2048,0.0754)
	(4096,0.0447)(8192,0.0305)(16384,0.0242)
	(32768,0.0170)
    };
    \addlegendentry{$ f_1$} 

\addplot[
    color=blue,
    mark=triangle*,
    ]
    coordinates {
    (512,.1120)(1024,0.0942)(2048,0.0810)
	(4096,0.0567)(8192,0.0365)(16384,0.0305)
	(32768,0.0249)
    };
    \addlegendentry{$ f_2 $} 
\addplot[
    color=black,
    mark=oplus*,
	style={densely dotted}
    ]
    coordinates {
    (512,.1002)(1024,.0840)(2048,0.0723)
	(4096,0.069)(8192,0.0592)(16384,0.0492)
	(32768,0.0300)
    };
    \addlegendentry{$ f_3 $} 
\end{axis}
\end{tikzpicture}\label{dd}}
	\caption{Evaluation of the estimate output by $ \SURF $ 
		with degrees $ d=1,2,3 $, $ \alpha=0.25$, on
		$ f_1 = 0.4\text{Beta}( 3, 4) + 0.6\text{Beta}( 5, 2)  $, 
		$ f_{2}= 0.4\text{Beta}( 10, 3)  + 0.6\text{Beta}( 2, 8)  $, and 
		$ f_{3} = \text{Beta}( 6, 6) $. }
	\label{plot:betasurf}
	\vspace{-1em}
\end{figure*}
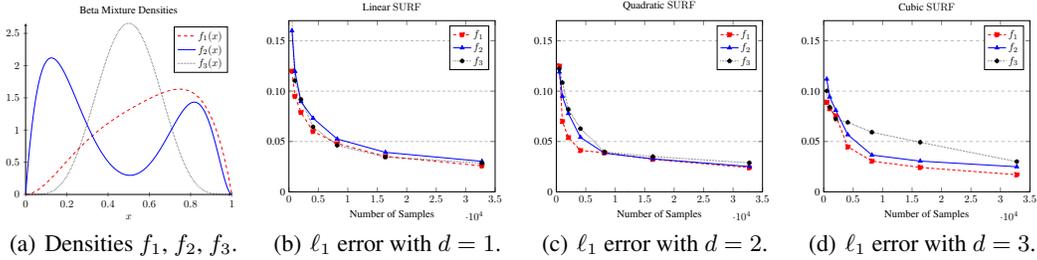
We now follow up with an experimental comparison.
$ \SURF $ is run with $ \alpha=0.25 $ and the 
errors are averaged over $ 10 $ runs. 
In running $ \ADLS $ we use the provided code as is.
Figure~\ref{plot:compare} compares the $ \ell_1 $
error in piecewise-linear estimation using $ \SURF $ vs 
$ \ADLS $ on the distributions considered 
in~\cite{jay17}, namely, a beta, Gamma, and 
Gaussian mixture. The plots correspond to the errors incurred on running $ \SURF $, 
and $ \ADLS $ with  pieces $ t=5,10,20,40,60 $.
While some hyperparameter 
optimizations may aid either algorithms, observe that 
the errors can be much larger 
with the wrong $t$. 
Significantly, the $ t=5 $ for which 
the results are comparable, is also roughly the 
number of pieces that $ \SURF $ outputs. 

Experiments show that $\SURF$ learns a wide range of 
parametric families such as the beta, 
Gaussian and Gamma distributions.
In Figure~\ref{plot:betasurf} we show results on 
beta mixture distributions
over $ [0,1] $, as they accommodate a wide range of shapes.
Other results may be found in Appendix~\ref{appen:experi}.
Let $ \text{Beta}(\alpha,\beta)$ be the beta density 
with parameters $ \alpha $, $ \beta $.
We run $ \SURF $ to estimate three distributions, as
shown in Figure~\ref{aa}.
$ \SURF $ estimates them using piecewise polynomials
of degree $ d=1,2,3 $. Figures~\ref{bb}--\ref{dd} show the 
resulting $ \ell_1 $ errors.
Observe that the errors are decaying, 
and are similar between distributions. 
This is not surprising since low degree polynomial approximations
largely rely on local smoothness, which all of the
considered densities possess. By the same reasoning,
on increasing $ d $ from $ 1 $ to $3 $,
the variation in error between distributions increases. 
The smoother $ f_1 $ starts incurring a smaller 
$ \ell_1 $ error than $ f_2 $ and $ f_3 $.
\begin{figure*}
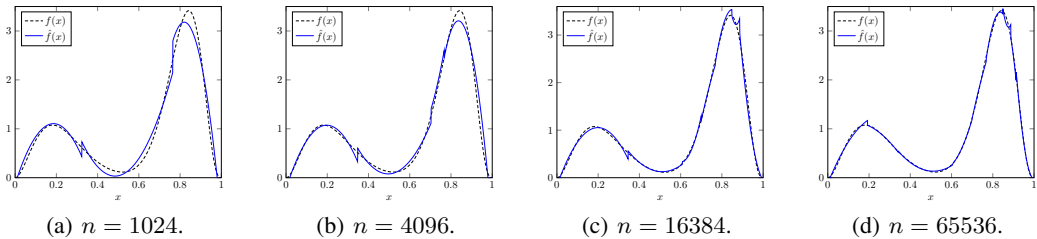

\centering
\subfigure[$ n=1024 $.]{\input{real1}}
\hfill
\subfigure[$ n=4096 $.]{\input{real4}}
\hfill
\subfigure[$ n=16384 $.]{\input{real16}}
\hfill
\subfigure[$ n=65536 $.]{\input{real64}}
\ignore{-1em}
\caption{$ \SURF $ with degree $ d=2 $, $ \alpha=0.25 $ 
estimating $ f= 0.3f_{\text{Beta}, 3, 10} 
+ 0.7f_{\text{Beta}, 17, 4} $ with $ n=1024,4096,16384, 65536 $ samples.}
\label{plot:action}
\ignore{-1.5em}
\end{figure*}
\begin{figure*}
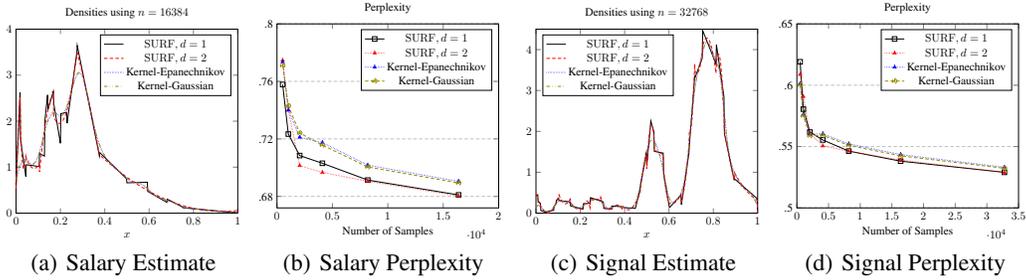

	\centering
	\subfigure[Salary Estimate]{\input{salplot}\label{ccc}}
	\hfill
	\subfigure[Salary Perplexity]{
\begin{tikzpicture}[scale=.43]
\begin{axis}[
title={Perplexity},
xlabel={Number of Samples},
xmin=0, xmax=20000,
ymin=0.64, ymax=.8,
xtick={0,5000,10000,15000,20000},
yticklabels={0,.64,.68,.72,.76,.8},
legend pos=north east,
ymajorgrids=true,
grid style=densely dashed,
]

\addplot[
color=black,
mark=square,
style=solid
]
coordinates {
(512, 0.7473)(1024,0.7042)(2048, 0.6855)
(4096,0.6788)(8192,0.6642)(16384, 0.6513)
};
\addlegendentry{$ \SURF, d=1$} 

\addplot[
color=red,
mark=triangle*,
style={densely dotted}
]
coordinates {
	(512, 0.7683)(1024,0.7250)(2048, 0.6768)
	(4096,.6708)(8192,0.6632)(16384,0.6504)
};
\addlegendentry{$ \SURF, d =2 $} 

\addplot[
 color=blue,
mark=triangle*,
style={densely dotted}
]
coordinates {
(512,0.7668)(1024,0.7250)(2048, 0.7014)
(4096,0.6968)(8192, 0.6772)(16384, 0.6631)
};
\addlegendentry{Kernel-Epanechnikov} 
\addplot[
color=olive,
mark=oplus,
style={densely dashed}
]
coordinates {
(512,0.7638)(1024,0.7287)(2048,0.7051)
(4096,0.6946)(8192,0.6756)(16384, 0.6617)
};
\addlegendentry{Kernel-Gaussian} 
\end{axis}
\end{tikzpicture}\label{ddd}}
	\hfill
	\subfigure[Signal Estimate]{\input{senplot}\label{aaa}}
	\hfill
	\subfigure[Signal Perplexity]{
\begin{tikzpicture}[scale=.43]
\begin{axis}[
title={Perplexity},
xlabel={Number of Samples},
xmin=0, xmax=35000,
ymin=0.5, ymax=.65,
xtick={0,5000,10000,15000,20000,25000,30000,35000},
yticklabels={0,.5,.55,.6,.65},
legend pos=north east,
ymajorgrids=true,
grid style=densely dashed,
]

\addplot[
color=black,
mark=square,
style=solid
]
coordinates {
(512, 0.6191)(1024,0.5805)(2048, 0.5618)
(4096,0.5554)(8192,0.5463)(16384,0.5381)(32768,0.5290)
};
\addlegendentry{$ \SURF, d=1$} 

\addplot[
color=red,
mark=triangle*,
style={densely dotted}
]
coordinates {
	(512, 0.6088)(1024,0.5908)(2048, 0.5609)
	(4096,.5505)(8192,0.5462)(16384,0.5390)(32768,0.5288)
};
\addlegendentry{$ \SURF, d =2 $} 

\addplot[
 color=blue,
mark=triangle*,
style={densely dotted}
]
coordinates {
(512,0.6013)(1024,0.5761)(2048, 0.5601)
(4096,0.5604)(8192, 0.5522)(16384,0.5435)(32768,0.5331)
};
\addlegendentry{Kernel-Epanechnikov} 
\addplot[
color=olive,
mark=oplus,
style={densely dashed}
]
coordinates {
(512,0.6002)(1024,0.5750)(2048,0.5588)
(4096,0.5591)(8192,0.5510)(16384,0.5424)(32768,0.5323)
};
\addlegendentry{Kernel-Gaussian} 
\end{axis}
\end{tikzpicture}\label{bbb}}
	\hfill
	\ignore{-1em}
	\caption{Real data estimates and
		perplexity of $ \SURF $ vs MLE based Kernel estimators}
	\vspace{-.5em}
	\label{fig:perp}
\end{figure*}

Next, we run $ \SURF $ with $ d=2$ to estimate 
$ f= 0.3f_{\text{Beta}, 3, 10} + 0.7f_{\text{Beta}, 17, 4} $ 
with $ n=1024,4096,16384, 65536 $. Figure~\ref{plot:action} plots 
the resulting estimates against $ f $. Notice that the estimate not only 
successively better estimates $ f $ in $ \ell_1 $ distance, 
but also pointwise converges to $f$.

Finally, we ran $\SURF$ on real data sets consisting of salaries from the 1994 US census and electric signals from the sensorless drive diagnosis dataset~\cite{dua19}, that have been used to evaluate classification algorithms~\cite{kohavi96,cheng99,pas13}.
We trim $0.5 \%$ of samples on either side and re-scale 
to obtain $ 57923 $ samples that lie in $ [0,1] $.
Figures~\ref{ccc} and \ref{aaa} show the estimate output 
by $ \SURF $ and the similarly non-parametric,
popularly used Kernel estimator with Epanechnikov 
and Gaussian kernels via the \verb|fitdist()|
function in \verb|MATLAB|\textsuperscript{\textregistered}.
As it can be observed, $\SURF$, without any hidden parameter,
recovers characteristic features of the distribution such as the
clusters, mode values, and tails. 
This is in contrast with $ \ADLS $, that, strictly speaking, cannot 
be used in this context as it requires additional cross-validation 
to tune $ t $ based on the number of clusters, etc.
The perplexity, or the exponent of the average 
negative log-likelihood on unseen
samples, is a commonly used measure in practice
to evaluate an estimate.
Figures~\ref{bbb},~\ref{ddd} compares the perplexity
on a test set with one-fourth the number of 
samples.
As it can be seen, even as 
\verb|fitdist()| outputs the perplexity minimizer on the 
training set, $ \SURF $ performs better.

\section*{Broader Impact}
SURF is a simple, universal, robust, and fast algorithm for the important problem of estimating distributions by piecewise polynomials. Real-life applications are likely to be approximated by relatively low-degree polynomials and require fast algorithms. SURF is particularly well-suited for these regimes. 
\section*{Acknowledgments}
We are grateful to the National Science Foundation (NSF) for supporting this work through grants CIF-1564355 and CIF-1619448.
\bibliography{ref}
\bibliographystyle{plain}

\onecolumn
\appendix
\section{Introduction}
\subsection{Proof of Theorem~\ref{thm:polyblack}}
\begin{Proof}
For a given $ d $, $\EQ$ outputs 
$\hatf_{I, \EQ}$, the re-scaled-shifted
$\hatf_{\barn_{d}}$ given by the corresponding
$ \barn_{d} \in \cN_d $ in Lemma~\ref{lem:optnodes}.
Choosing {$\epsilon(\delta)= \sqrt{{5\log({1}/{\delta})}/{n}}$, for $n\ge 128$,
$ \cQ_{\epsilon(\delta)} $} occurs with probability 
$ \ge 1-\delta $ from Lemma~\ref{lem:multconc}.
Using Lemma~\ref{lem:ratioconst} with $ \epsilon(\delta) $ completes the proof.
\end{Proof}

\subsection{Proof of Theorem~\ref{thm:interhistopt}}
\begin{Proof}
	Choosing $\epsilon(\delta)= \sqrt{{5\log({n}/{\delta})}/{n}}$, 
	for $n\ge 128$, $ \cQ_{\epsilon(\delta)} $ occurs with probability
	$ \ge 1-\delta $ from Lemma~\ref{lem:multconc}.
	Using Lemma~\ref{lem:binarycomp} 
	on top of Lemma~\ref{lem:master} proves the theorem.
\end{Proof}

\subsection{Proof of Corollary~\ref{cor:interhistopt}}
\begin{Proof}
	From Theorem~\ref{thm:interhistopt}, 
	w.p. $ \ge 1-\delta $,
	\begin{align*}
	\lone{\hatf_{\cA(X^{n-1})}-f}
	&\le
	\min_{\barI\in \Delta_{\reals}(X^{n-1})}\sum_{I \in \barI} 
	\left(
	\frac{(r_d+1)\cdot \alpha}{\alpha-2}\cdot 
	\inf_{\poly \in \cP_d} \loneint{I}{\poly - f }\right. \\&\phantom{\min_{\barI\in \Delta_{\reals}}\sum_{I \in \barI}}
	\quad \quad
	+\left.\frac{r_d\cdot (\alpha\sqrt2+\sqrt{2}-1)}{(\sqrt{2}-1)^2}
	\sqrt{\frac{5(d+1) \hatp_I \log \frac{n}{\delta}}{n}}\right)\\
	&\overset{(a)}{\le}\min_{t\ge 0}
	\left(\frac{(r_d+1)\cdot\alpha}{\alpha-2} \mathrm{OPT}_{\cP_{t,d}} +\frac{r_d\cdot(\alpha\sqrt2+\sqrt{2}-1)}{(\sqrt{2}-1)^2}
	\sqrt{\frac{5t\cdot (d+1) \log \frac{n}{\delta}}{n}}\right)\\
	&\overset{(b)}{\le}\min_{t\ge 0}
	\left( 
(r_d+1)\left(1+\frac{4}{\alpha}\right)\cdot \mathrm{OPT}_{\cP_{t,d}}(f)\right.\\
&\phantom{\min_{t\ge 0}}\left.
+\frac{r_d\cdot(\alpha\sqrt2+\sqrt{2}-1)}{(\sqrt{2}-1)^2}
	\sqrt{\frac{5t\cdot (d+1) \log \frac{n}{\delta}}{n}}\right),
	\end{align*}
	where $ (a) $ follows since for any partition with $ t $ pieces, 
	$ \sum_{I\in \barI} \sqrt{\hatp_I} \le \sqrt{t}$,
	and $(b)$ follows since for any $x>4$ 
	$x/(x-2)< 1+4/x$.
	Letting $ \alpha\rightarrow \infty $
	and choosing $ \delta \approx 1/n $ completes the proof.
\end{Proof}\label{appen:main}
\section{Intervals and Partitions}
\subsection{Proof of Lemma~\ref{lem:singleconc}}
\begin{Proof}
For simplicity, let 
$$ X\ed \intp_{a,b} ,\ \ p \ed  \hatp_{a,b}$$ so that
$ X = \intp_{a,b}\sim \Beta(n \hatp_{a,b},
n(1-\hatp_{a,b})) = \Beta(np,n(1-p))$.
For any $x,y\in\reals^+$, let
$\text{B}(x,y)=\Gamma(x)\Gamma(y)/\Gamma(x+y)$
denote the beta function and let
$a,b>0$ and $x\in[0,1]$, 
$$I_x(a,b) \ed \int_{0}^{x} 
\frac{z^{a-1}(1-z)^{b-1}}{\text{B}(a,b)} dz$$
be the incomplete beta function. Then,
\begin{align*}
\Pr[X\le p-\epsilon\sqrt{p}]&\overset{(a)}{=} 
I_{p-\epsilon\sqrt{p}}(np,n(1-p))\\
&\overset{(b)}{=}\sum_{i=np}^{n-1} {n-1\choose i} 
(p-\epsilon\sqrt{p})^i(1-p+\epsilon\sqrt{p})^{n-1-i}\\
&\overset{(c)}{\le} e^{-(n-1)D(p||p-\epsilon\sqrt{p})}\\
&\overset{(d)}{\le} e^{-(n-1)\frac{\epsilon^2}{2}},
\end{align*}
where $(a)$ follows by definition,
$(b)$ follows by the property of 
incomplete beta function~\cite{incompnist}, 
$(c)$ follows from the Chernoff bound 
applied to the right tail of a $\text{Binom}(n,p-\epsilon\sqrt{p})$ 
random variable, and $ (d) $ follows since 
$ D(x||y)\le (x-y)^2/\max\{x,y\} $. Similarly,
\begin{align*}
\Pr[X\ge p+\epsilon\sqrt{p}]
&\overset{(a)}{=}1-I_{(p+\epsilon\sqrt{p})}(np,n(1-p))\\
&\overset{(b)}{=}1-
\sum_{i=np}^{n-1} {n-1\choose i} 
(p+\epsilon\sqrt{p})^i(1-p-\epsilon\sqrt{p})^{n-1-i}\\
&\le\sum_{i=0}^{np} {n-1\choose i} 
(p+\epsilon\sqrt{p})^i(1-p-\epsilon\sqrt{p})^{n-1-i}\\
&\overset{(c)}{\le} e^{-(n-1)D(p||p+\epsilon\sqrt{p})}\\
&\overset{(d)}{\le} e^{-(n-1)\frac{\epsilon^2 p}{2(p+\epsilon\sqrt{p})}},
\end{align*}
where $(a)$ follows by definition,
$(b)$ follows by the property of 
incomplete beta function~\cite{incompnist}, and
$(c)$ follows from Chernoff bound applied to the left tail of a
$\text{Binom}(n,p+\epsilon\sqrt{p})$ random variable, and 
$ (d) $ follows since $ D(x||y)\le (x-y)^2/\max\{x,y\} $.
\end{Proof}

\subsection{Proof of Lemma~\ref{lem:multconc}}
\begin{Proof}
From using the union bound, we have
\begin{align*}
1-\Pr[\cQ_{\epsilon}]&=
\Pr\left[\exists 0\le a<b\le n :
|\intp_{a,b}-\hatp_{a,b}|\ge
\epsilon \sqrt{\hatp_{a,b}}\right]\\
&\le \sum_{0\le a<b\le n }
\Pr\left[
|\intp_{a,b}-\hatp_{a,b}|\ge
\epsilon\sqrt{\hatp_{a,b}}\right]\\
&\overset{(a)}{\le} \sum_{0\le a<b\le n }
\left(e^{\frac{-(n-1)\epsilon^2}{2}}
+e^{\frac{-(n-1)\epsilon^2 \hatp_{a,b}}
{2(\hatp_{a,b}+\epsilon\sqrt{\hatp_{a,b}})}}\right)\\
&\overset{(b)}{=}\frac{n(n+1)}{2}
\left(e^{\frac{-(n-1)\epsilon^2}{2}}
+e^{\frac{-(n-1)\epsilon^2 \hatp_{a,b}}
{2(\hatp_{a,b}+\epsilon\sqrt{\hatp_{a,b}})}}\right)\\
&\overset{(c)}{\le} \frac{n(n+1)}{2}
\left(e^{\frac{-(n-1)\epsilon^2}{2}}
+e^{-\frac{(n-1)\epsilon^2
}{2(1+\epsilon\sqrt{n})}}\right),
\end{align*}
where $(a)$ follows from Lemma~\ref{lem:singleconc},
$(b)$ follows since $ |0\le a<b\le n| = 
\binom{n+1}{2} $, and $ (c) $ follows since $q_{a,b}\ge 1/n $. 
\end{Proof}\label{appen:conc}
\section{The Interpolation Routine}
\subsection{Proof of Lemma~\ref{lem:ratioconst}}
\begin{Proof}
For a partition $ \bar{I}$ of $ I=[0,1] $,
and integrable functions $ g_1, g_2 $, define the distance 
\begin{equation}
\label{eqn:ddistance}
d_{\bar{I}} (g_1,g_2) 
\ed \sum_{J\in \bar{I}} \Big|\int_J g_1 - \int_J g_2\Big|.
\end{equation}
In words, $ d_{\bar{I}} (g_1,g_2)  $ is the sum of
absolute differences between measures under
$ g_1 $ and $ g_2 $ across all intervals in $ \bar{I} $.
For any $ \poly \in \cP_d $,
\begin{align*}
\loneint{I}{\hatf_{\barn_d}-f} &\le \loneint{I}{\poly-f} + 
\loneint{I}{\hatf_{\barn_d}-\poly} \\
&\overset{(a)}{\le} \loneint{I}{\poly-f} + r_d(\barn_d)d_{\bar{I}_{\barn_d}}(\poly,\hatf_{\barn_d})\\
&\overset{(b)}{\le} \loneint{I}{\poly-f} + 
r_d(\barn_d)\left(d_{\bar{I}_{\barn_d}}(\poly,f) + d_{\bar{I}_{\barn_d}}(f,\hatf_{\barn_d})\right)\\
&\overset{(c)}{\le} \left(1+r_d(\barn_d)\right)\loneint{I}{\poly-f}+ 
r_d(\barn_d)d_{\bar{I}_{\barn_d}}(f,\hatf_{\barn_d})\\
&\overset{(d)}{\le} \left(1+r_d(\barn_d)\right)\loneint{I}{\poly-f} + 
r_d(\barn_d)\sum_{J\in \barI_{\barn_d}}
|\intp_J-\hatp_J|.
\end{align*}
where $ (a) $ follows since $ (\poly-\hatf_{I,\barn_d})\in \cP_d $,
and from definitions of the ratio $ r_d(\barn_d) $ 
in Equation~\eqref{eqn:polyrat},
$ (b) $ follows since the $ d_{\barI} $-distance 
satisfies the triangle inequality,
$ (c) $ follows since the $ \ell_1 $ distance upper
bounds $ d_{\barI}-$distance, 
$ (d) $ follows since $ \hatf_{\barn_d} $,
by definition, is the polynomial such that
$ \int_J \hatf_{\barn_d} = \hatp_{J}$ $ \forall J\in \barI_{\barn_{d}} $, and 
the interval probability $ \intp_{J}\ed \int_{J} f $.

Since $ \intp_J\sim \text{Beta}(\hatp_{j}n, (1-\intp_{J})n) $,
it follows that 
\begin{align*}
\EE \loneint{I}{\hatf_{\barn_{d}}-f}&\le 
(1-r_{d}(\barn_{d}))\loneint{I}{h-f}+r_{d}(\barn_{d})\sum_{J\in \barI_{\barn_d}}
\EE|\intp_J-\hatp_J|\\
&\overset{(a)}{\le}(1-r_{d}(\barn_{d}))\loneint{I}{h-f}+r_{d}(\barn_{d})\sum_{J\in \barI_{\barn_d}}
\sqrt{\frac{2q_J}{\pi n}}\\
&\overset{(b)}{\le} \left(1+r_d(\barn_d)\right)\loneint{I}{\poly-f} + 
r_d(\barn_d)
\sqrt{\frac{2(d+1)\hatp_{I}}{\pi n}}.
\end{align*}
where $ (a) $ follows from the mean absolute deviation of the 
Beta distribution, and $ (b) $ follows since by the concavity of 
$ \sqrt{x} $ for $ x\ge 0 $, the sum 
$ \sum_{J\in \barI_{\barn_d}} \sqrt{ \hatp_J } $ is
maximized if for each $ J\in \barI_{\barn_{d}} $, 
$\hatp_J = (\sum_{J\in 
\barI_{\barn_d}} \hatp_J)/|\barI_{\barn_d}|= \hatp_{I}/(d+1)$.

The following version will be useful in the proof of Lemma~\ref{lem:master}.
Under event $ \cQ_{\epsilon}$,
\begin{align*}
\EE \loneint{I}{\hatf_{\barn_{d}}-f}&\le 
(1-r_{d}(\barn_{d}))\loneint{I}{h-f}+r_{d}(\barn_{d})\sum_{J\in \barI_{\barn_d}}
\EE|\intp_J-\hatp_J|\\
&\le 
\left(1+r_d(\barn_d)\right)\loneint{I}{\poly-f} + 
r_d(\barn_d)\sum_{J\in \barI_{\barn_d}}
\epsilon\sqrt{ \hatp_J }\\
&\overset{(a)}{\le} \left(1+r_d(\barn_d)\right)\loneint{I}{\poly-f} + 
r_d(\barn_d)
\epsilon\sqrt{(d+1)\hatp_I },
\end{align*}
where $ (a) $ follows due to the same reasoning as above.
\end{Proof}

\subsection{Proof of Lemma~\ref{lem:zeroish}}
\begin{Proof}
Fix $ \poly\in \cP_{d} $. Let $(\beta_1,\upto ,\beta_{d_0}) $ be
the roots of $ \poly$ in $ [0,1] $ for some 
$ \beta_1\le \upto \le \beta_{d_{0}}, 0\le d_{0}\le d $.
Let $ \beta_{0}\ed 0, \beta_{d_{0}+1}\ed 1 $.
Notice that 
\begin{align*}
\int_{0}^{1} |h|  =\sum_{i=1}^{d_0+1}\Big |\int_{\beta_{i-1}}^{\beta_{i}}  h\Big |
&\overset{(a)}{\le} \sup_{\barm_d\in \cN_d } 
\sum_{i=1}^{d+1}\Big |\int_{m_{i-1}}^{m_{i}}  h\Big |
\\&=\sup_{\barm_d\in \cN_d } \max_{\bar{s}\in \{0,1\}^{d+1} }
\sum_{i=1}^{d+1}(-1)^{s_i}\int_{m_{i-1}}^{m_{i}}  h\\
&\overset{(b)}{\le}\int_{0}^{1} |h|,
\end{align*}
where $ (a) $ follows since on padding $ d-d_{0} $ zeros,
$ (0,\cdots, 0, \beta_{0}\upto \beta_{d+1})\in \cN_d $.
Thus $ (b) $ is, in fact, an equality, implying
\begin{align*}
r_{d}(\barn_d) 
=\sup_{\poly\in \cP_d}r(\barn_{d},\poly)
&= \sup_{\poly\in \cP_d}
\frac{\int_{0}^1|h|}{\sum_{i=1}^{d+1}|\int_{n_{i-1}}^{n_i}
\poly|}\\
&=
\sup_{\poly\in \cP_d}
\frac{\sup_{\barm_d\in \cN_d}\max_{\bar{s}\in \{0,1\}^{d+1} }
\sum_{i=1}^{d+1}(-1)^{s_i}\int_{m_{i-1}}^{m_i}h}
{\sum_{i=1}^{d+1}|\int_{n_{i-1}}^{n_i}h|}\\
&=\sup_{\barm_d\in \cN_d}\max_{\bar{s}\in \{0,1\}^{d+1} }
\sup_{\poly\in \cP_d}
\frac{\sum_{i=1}^{d+1}(-1)^{s_i}\int_{m_{i-1}}^{m_i}h}
{\sum_{i=1}^{d+1}|\int_{n_{i-1}}^{n_i}h|}.
\end{align*}
Denote $ h = \sum_{i=1}^{d+1} c_{i}\cdot  x^{i-1}  $
and let $ \bar{c}\ed(c_1\upto c_{d+1}) $.
Notice that since $ r(\barn_{d}, \poly)\ge 1 $
for any $ \poly \in \cP_d $, and since $ r_d(\barn_{d}, 0)\ed 1 $,
WLOG assume $ \poly \neq 0 $ or
$ \bar{c}\neq \bar{0}\ed (0\upto 0) $.
By linearity of the integral of $\poly $ in $ \bar{c}$, recast $ r_{d}(\barn_{d}) $ into
\[
r_{d}(\barn_{d}) = 
\sup_{\barm_{d}\in {\cN_{d}}}\max_{\bar{s}\in \{0,1\}^{d+1} }
\sup_{\bar{c}\in \reals^{d+1}\setminus\{\bar{0}\}}
\frac{\sum_{i=1}^{d+1}c_i\mu_{i}}
{\sum_{i=1}^{d+1}|\sum_{j=1}^{d+1}c_j\lambda_{i,j}|},
\]
where for any $ i, j \in \{1\upto d+1\} $,
$ \mu_{i}\in \reals $ is a function of $ \barm_d,\bar{s} $ and
$\lambda_{i,j} \in \reals$ is a function of $\barn_d $.
Observe that $ \barn_{d} $ is given, and additionally 
fix $ \barm_{d}\in \cN_{d} $, $ \bar{s}\in \{0,1\}^{d+1} $.
Since the objective function here is a ratio whose denominator is 
positive (since $ h\neq 0 $), WLOG set the numerator 
to $ 1 $ via the constraint $\sum_{i=1}^{d+1}c_i\mu_{i}=1 $ 
and convert it to a linear program as:
\[
\max
\frac{1}{\sum_{i=1}^{d+1} v_i } : \bar{c}, \bar{v} \in \reals^{d+1}, \
\ v_i\ge \sum_{j=1}^{d+1}c_j\lambda_{i,j}, \
v_i \ge -\sum_{j=1}^{d+1}c_j\lambda_{i,j},
\sum_{i=1}^{d+1} c_i\mu_i =1,
\]
where $ \bar{v}\ed (v_1\upto v_{d+1})  $.
Observe that these constraints give rise to a 
bounded region, and since this is a linear program, 
there exists a solution at some corner point 
involving at least $ 2\cdot (d+1) $ equalities, 
one for each variable.
In any such solution, since the equality:
$ \sum_{i=1}^{d+1} c_i \mu_i =1 $ is always active,
at least $ 2\cdot (d+1) - 1 $ of the other 
inequalities attain equality. 
Notice that for any $ i\in\{1\upto d+1\} $, 
$ v_i=0 $ if both
\[  v_i = \sum_{j=1}^{d+1}c_j\lambda_{i,j} \text{ and } 
v_i =  -\sum_{j=1}^{d+1}c_j\lambda_{i,j} \text{ hold}.\] 
Thus in this corner point solution, $ v_i \neq 0 $ for at most one $ i\in \{1\upto d+1\} $.
Let 
\[ \cD_{\barn_{d}} = \Big \{\bar{c} \in \reals^{d+1} \setminus \{\bar{0}\}:
\exists i_1\in \{1\upto d+1\}:
\forall i\neq i_1, \ |\sum_{j=1}^{d+1}c_j\lambda_{i,j}|=0\Big \}\]
This implies
\begin{align*}
r_{d}(\barn_d) 
&=\sup_{\barm_{d}\in {\cN_{d}}}\max_{\bar{s}\in \{0,1\}^{d+1} }
\sup_{\bar{c}\in \reals^{d+1}\setminus \{\bar{0}\}}
\frac{\sum_{i=1}^{d+1}c_i\mu_i}
{\sum_{i=1}^{d+1}|\sum_{j=1}^{d+1}c_j\lambda_{i,j}|}\\
&=\sup_{\barm_{d}\in {\cN_{d}}}\max_{\bar{s}\in \{0,1\}^{d+1} }
\max_{\bar{c}\in \cD_{\barn_{d}}}
\frac{\sum_{i=1}^{d+1}c_i\mu_i}
{\sum_{i=1}^{d+1}|\sum_{j=1}^{d+1}c_j\lambda_{i,j}|}\\
&=\sup_{\barm_{d}\in {\cN_{d}}}\max_{\bar{s}\in \{0,1\}^{d+1} }
\max_{\poly\in \cP_{\barn_d}}
\frac{\sum_{i=1}^{d+1}(-1)^{s_i}\int_{m_{i-1}}^{m_i}h}
{\sum_{i=1}^{d+1}|\int_{n_{i-1}}^{n_i}h|}\\
&= \max_{\poly\in \cP_{\barn_d}}
\sup_{\barm_{d}\in {\cN_{d}}}\max_{\bar{s}\in \{0,1\}^{d+1} }
\frac{\sum_{i=1}^{d+1}(-1)^{s_i}\int_{m_{i-1}}^{m_i}h}
{\sum_{i=1}^{d+1}|\int_{n_{i-1}}^{n_i}h|}\\
&= \max_{\poly\in \cP_{\barn_d}}
\frac{\int_{0}^{1}|h|}
{\sum_{i=1}^{d+1}|\int_{n_{i-1}}^{n_i}h|}
= \max_{\poly\in \cP_{\barn_d}} r_{d}(\barn_d,h).
\end{align*}
\end{Proof}

\subsection{Proof of Lemma~\ref{lem:optnodes}}
\begin{Proof}
For any polynomial $ \poly\in \cP_{d} $, the ratio
$r(\barn_d, \poly) $ is invariant to
multiplying both the numerator and
denominator by a constant. Thus, WLOG consider
polynomials whose leading coefficient is 1. 
Then for any $ \barn_d\in \cN_d $, $ \cP_{\barn_d} 
= (\poly_{\barn_d,1}\upto \poly_{\barn_d,d+1})$, 
is a set consisting of $ d+1 $ unique polynomials, where
each $ \poly_{\barn_d,i}, \ i\in \{1\upto d+1\} $ 
is that polynomial with 0 area in all intervals 
in $ \barI_{\barn_d} $ except $ I_{\barn_d,i}$.

\textbf{Case $ \mathbf{d = 0} $:} 
Here $ \cN_0=\{(0,1)\} $ and is a singleton set. 
Since any $ h\in\cP_0$ is a constant
value, $ \int_0^{1} |h| = |\int_0^{1} h |$. Therefore
$ r_0^{\star}=\max_{\poly \in \cP_0}
r(\barn_{d},h) = 1$.

\textbf{Case $ \mathbf{d = 1} $:} Let $ \barn_1=(0,m,1) $.
In this case $ \poly_{\barn_d,1}(x) = x-m/2  $
and $ \poly_{\barn_d,2}(x) = x-(1+m)/2  $.
Using Lemma~\ref{lem:zeroish}, 
\begin{align*}
r_1(\barn_d) 
&= \max_{\poly \in \cP_{\barn_d}}r(\barn_d,h)
= \max\{r(\barn_d, \poly_{\barn_d,1}),
r(\barn_d,\poly_{\barn_d,2})\}\\
&= \max\Big\{ \frac{m^2/4+(1-m/2)^{2}}{1-m},
\frac{(1-m)^2/4+((1+m)/2)^{2}}{m}\Big\}.
\end{align*}
$ r_1(\barn_d)  $ is minimized for $ m^{\star} = 1/2 $,
giving $ r^{\star}_1=(1/16+9/16)/(1/2)=1.25 $.

\textbf{Case $ \mathbf{d = 2} $:}
By symmetry, the minimizing node partition
is symmetric about $ 0.5 $. Thus WLOG let $ \barn_2 =(0,m,1-m,1) $ for 
some $ m\le 0.5 $.  Among the $ d+1=3 $ polynomials 
in $ \cP_{\barn_2} $, by symmetry of $ \barn_2 $, 
$ r_{2}(\poly_{\barn_d,1})  =r_{2}(\poly_{\barn_d,3})$. 
Thus we consider the larger ratio across only two polynomials,
$ \poly_{\barn_d,2}, \poly_{\barn_d,3} $.

Denote the polynomial as $ \poly_{\barn_d,2}(x) =
(x-a_{2})^2-b_{2}^2 $ and upon setting the respective integrals to $ 0 $,
\[\Big|\frac{m^3}{3}-a_{2}m^2+(a_{2}^2-b_{2}^2)m\Big|
=0, \ \Big|\frac{1-(1-m)^3}{3}-a_{2}(1-(1-m)^2)
+(a_{2}^2-b_{2}^2)m\Big|=0
\]
\[\implies a_{2}=\frac12, b_{2}^{2}=\frac{3(m^2-m)+1}{9}.\]
Representing $ \poly_{\barn_d,3}(x) = (x-a_{3})^2-b_{3}^2 $
and repeating the same steps,
\[
\Big|\frac{m^3}{3}-a_{3}m^2+(a_{3}^2-b_{3}^2)m\Big|
=0,\ \Big|\frac{(1-m)^3-m^3}{3}-a_{3}((1-m)^2-m^2)
+(a_{3}^2-b_{3}^2)(1-2m)\Big|=0
\]
\[\implies a_{3}=\frac13, b_{3}^{2}=\frac{4m^2-6m+3}{3}.\]

The corresponding $ r(\barn_d,\poly_{\barn_d,2}) $ and 
$ r(\barn_d,\poly_{\barn_d,3}) $ are given by
\[\frac{8\left(\frac{1-3m(1-m)}{9}\right)^{3/2}}{m(1-m)}+1,
\,
\frac{2\left(\frac{(2m-1)(2m-2)+1}{3}\right)^{3/2}}{(2m-1)(m-1)}-1. \]

From simultaneously minimizing the above expressions by equating 
them, the optimal $  m $ is the root of 
\begin{align*}
q_{2}(m)&=-\frac{26624}{729} m^{14}+\frac{193280}{729}m^{13}
- \frac{211024} {243} m^{12} + \frac{3703648}{2187} m^{11} 
- \frac{4790776}{2187} m^{10}
\\
&\phantom{=}
+ \frac{39108232}{19683} m^9  
-\frac{8554775}{6561} m^8 
+\frac{12357280}{19683} m^7 
- \frac{13004032}{59049} m^6 
+ \frac{1061792}{19683} m^5 
\\
&\phantom{=}
- \frac{4350752}{531441} m^4 
+ \frac{246976}{531441}m^3 
+ \frac{11840}{177147} m^2 
- \frac{6656}{531441}m 
+ \frac{256}{531441}
\end{align*}
near $ 0.26 $. Thus the optimal $m^{\star}\approx 0.2599$
and the corresponding $r_2^{\star}\approx 1.423$.

\textbf{Case $ \mathbf{d = 3} $:}
By symmetry, as before, WLOG let $ \barn_d=(0,m,0.5,1-m,1) $.
This reduces the search space to just two polynomials, 
$ \poly_{\barn_d,1} $, $\poly_{\barn_d,2}  $. 
The optimal $ m$ occurs as the root of
\begingroup
\allowdisplaybreaks
\begin{align*}
q_3(m)&\ed m^{69}
+\frac{2233}{46}m^{68}
+\frac{3394851}{2944}m^{67}
-\frac{26295551}{1472}m^{66}
+\frac{76466381715}{376832}m^{65}
\\&\phantom{=}
-\frac{1357944230009}{753664}m^{64}
+\frac{627961733592749}{48234496}m^{63}
-\frac{3795194179761079}{48234496}m^{62}
\\&\phantom{=}
+\frac{1252499739594399621}{3087007744}m^{61}
-\frac{5593584650474780121}{3087007744}m^{60}
\\&\phantom{=}+\frac{87541700408454835933}{12348030976}m^{59}
-\frac{9689649149944354300097}{395136991232}m^{58}
\\&\phantom{=}+\frac{477481388280085878102175}{6322191859712}m^{57}
-\frac{1316816736336377796401265}{6322191859712}m^{56}\\&\phantom{=}+\frac{104518853645525535726426411}{202310139510784}m^{55}
-\frac{935729957191660731330480575}{809240558043136}m^{54}\\
&+\frac{1894032003216065918256250147}{809240558043136}m^{53}
-\frac{111070063686665905121657252873}{25895697857380352}m^{52}\\
&+\frac{2947966937880382636398337723253}{414331165718085632}m^{51}
-\frac{96053130159826779148472826511}{9007199254740992}m^{50}\\
&+\frac{5957768773291898355944143881565}{414331165718085632}m^{49}
\\&
-\frac{28666092800309188568756667723285}{1657324662872342528}m^{48}\\
&+\frac{1936825259310147713677259087614429}{106068778423829921792}m^{47}
\\
&
-\frac{216663456959495677102483903955187}{13258597302978740224}m^{46}\\
&+\frac{4797689934446961630160031643189779}{424275113695319687168}m^{45}
\\&
-\frac{6609917603386978813128872815736861}{1697100454781278748672}m^{44}\\
&-\frac{1831051294952734349349124229564767}{424275113695319687168}m^{43}
\\&
+\frac{77448257249275962672807384094624197}{6788401819125114994688}m^{42}\\
&-\frac{3438211574596414864435399648557575571}{217228858212003679830016}m^{41}
\\&
+\frac{14691873341415043555417961121466375911}{868915432848014719320064}m^{40}\\
&-\frac{52197058412928213666930233026164438477}{3475661731392058877280256}m^{39}\\&
+\frac{626588406181557032617836659688444683449}{55610587702272942036484096}m^{38}\\
&-\frac{1555525051509188771980730278198868813547}{222442350809091768145936384}m^{37}\\&
+\frac{2911728348738530370986950039544396475929}{889769403236367072583745536}m^{36}\\
&-\frac{415629763783269606797247480824967018319}{618970019642690137449562112}m^{35}\\&
-\frac{21293950227855325096203381076755307029285}{28472620903563746322679857152}m^{34}\\
&+\frac{568117182824342053622453013472967651433283}{455561934457019941162877714432}m^{33}\\&
-\frac{544176155073471826876603760623685652791461}{455561934457019941162877714432}m^{32}\\
&+\frac{26510792231823207111468208052472537291091795}{29155963805249276234424173723648}m^{31}\\
&-\frac{4339252967361049353363803334422356143484711}{7288990951312319058606043430912}m^{30}\\
&+\frac{80732188658524254663695919868749449919506951}{233247710441994209875393389789184}m^{29}\\
&-\frac{21178612540977853104278860543475186392271681}{116623855220997104937696694894592}m^{28}\\
&+\frac{80945977995703186772569013525777965350114711}{932990841767976839501573559156736}m^{27}\\
&-\frac{17702912107607481296775724392303230443070095}{466495420883988419750786779578368}m^{26}\\
&+\frac{113795689628398207709330045266274999410341749}{7463926734143814716012588473253888}m^{25}\\
&-\frac{21034631455622622763256522489198166970683613}{3731963367071907358006294236626944}m^{24}\\
&+\frac{229167532475013797669274973054071362594872031}{119422827746301035456201415572062208}m^{23}\\
&-\frac{17963282649430914596699742603270996080670419}{29855706936575258864050353893015552}m^{22}\\
&+\frac{165857532854257459359651249063603517430127219}{955382621970408283649611324576497664}m^{21}\\
&-\frac{21988393007280087042133518345726499678514793}{477691310985204141824805662288248832}m^{20}\\
&+\frac{2673062517068208952962202401457887091620815}{238845655492602070912402831144124416}m^{19}\\
&-\frac{4753573760248527366859497435267453417384687}{1910765243940816567299222649152995328}m^{18}\\
&+\frac{7698530432612172216840174126420917394335697}{15286121951526532538393781193223962624}m^{17}
\\&
-\frac{88254391793832097376072061503233796540539}{955382621970408283649611324576497664}m^{16}\\
&+\frac{466181759012121800363517839555926856292931}{30572243903053065076787562386447925248}m^{15}\\&
-\frac{34314489379863699139383468530966229926169}{15286121951526532538393781193223962624}m^{14}\\
&+\frac{71176987410160907949890583502121075660389}{244577951224424520614300499091583401984}m^{13}\\&
-\frac{3987333962073343668163889901656755306225}{122288975612212260307150249545791700992}m^{12}\\
&+\frac{748668655114740354600745066271570265031}{244577951224424520614300499091583401984}m^{11}\\&
-\frac{110845256761306163606440669442292637027}{489155902448849041228600998183166803968}m^{10}\\
&+\frac{1899176964083859703283044829743633209}{170141183460469231731687303715884105728}m^{9}\\&
-\frac{2073921584792354563737120211683341}{42535295865117307932921825928971026432}m^{8}\\
&-\frac{3440239020182100263379082512521175}{61144487806106130153575124772895850496}m^{7}
\\&+\frac{4264138856148752641548430451717}{664613997892457936451903530140172288}m^{6}\\
&-\frac{3117503551035781118929644883731}{7643060975763266269196890596611981312}m^{5}\\&
+\frac{29037077182037119112722125423}{1910765243940816567299222649152995328}m^{4}\\
&-\frac{46083309361423573178372679}{238845655492602070912402831144124416}m^{3}\\&
-\frac{597811927318403605685541}{59711413873150517728100707786031104}m^{2}\\
&+\frac{30104861649982869480831}{59711413873150517728100707786031104}m\\&
-\frac{105905655782897976459}{14927853468287629432025176946507776}
\end{align*}
\endgroup
 
near $ 0.155 $. This gives $ m^{\star}\approx 0.1548 $
and the corresponding $ r^{\star}_{3}\approx 1.559 $.

For degrees $ 4\le d\le 8 $, we use numerical
methods on top of Lemma~\ref{lem:zeroish} to derive 
$ \barn_d\in \cN_d $ and the corresponding $ r_d(\barn_d) $.
These values populate the second table in Lemma~\ref{lem:optnodes}.
\end{Proof}
\label{appen:poly}
\section{The Compare and Stitch Routines}
\subsection{Proof of Lemma~\ref{lem:binarycomp}}
\begin{Proof}
Observe that 
any $ \hatp\in \barhatp\in \Delta_{\textrm{emp,n}} $
is an integral multiple of $ 1/n $. As $ \log _{2}n $ 
is an integer, we may decompose $ \hatp $ along its 
binary expansion as \[ \hatp = \sum_{j=0}^{\log_{2} n} 2^{-j}b_{j},\]
for some $ b_{j}\in\{0,1\} $, $ j\in \{1\upto \log_{2}n\} $. 
Replace each $ \hatp \in \barhatp$ with the vector 
$ (2^{-0}b_{0}, 2^{-1}b_{1},\cdots) $ to obtain 
$\barhatp'\in \Delta_{\mathrm{bin,n}}$.
From the property of the geometric sum,
\[\sum_{j=0}^{\log_{2}n} \sqrt{2^{-j}b_{j}}
\overset{}{\le}
\frac{\sqrt{\hatp}}{\sqrt2-1}.\]
Finally $\lone{\starf_{\barI_{\barhatp'}}-f} 
\le
\lone{\starf_{\barI_{\barhatp}}-f}$
since $\barI_{\barhatp'}$ being a finer partition
than $ \barI_{\barhatp} $, 
$ \starf_{\barI_{\barhatp'}} $ is a closer approximation to $ f $ 
than $ \starf_{\barI_{\barhatp}} $.
\end{Proof}

\subsection{Proof of Lemma~\ref{lem:master}}
\begin{Proof}
For any interval $ I $, let 
\[ f^{\star}_{I} \ed \arg \min_{h\in \cP_{d}}\loneint{I}{h-f},\]
and for any partition $ \barI $, let $  f^{\star}_{\barI}  $
be the piecewise polynomial that equals 
$ f^{\star}_{I} $ in each $ I\in \barI $.
For simplicity let 
$ I_{\barhatp}\ed I_{\barhatp_D} $ denote the final
partition and $ \barhatp \ed \barhatp_{D} $ the 
corresponding empirical distribution. Consider any 
$ \barhatpp \in \Delta_{\mathrm{bin,n}}$ 
and its associated interval partition, 
$ \barI_{\barhatpp} $.
Two interval partitions $ \bar{I}_1,\bar{I}_2 $ 
corresponding to binary distributions 
have the following property:
Any interval in $ \bar{I}_1 $ is either 
completely contained within some interval in $ \bar{I}_2 $, 
or is a union of contiguous intervals from $ \bar{I}_2 $.
As a result $ \barI_{\barhatp} $ may partitioned 
into three classes of intervals:
\begin{figure}[H]
\centering
\includegraphics[scale=0.7]{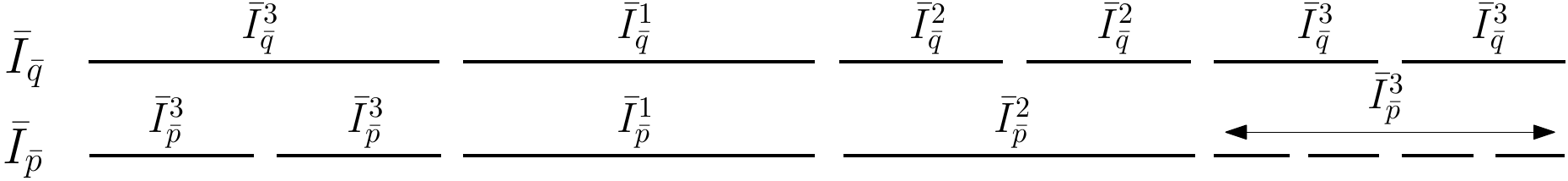}
\caption{Illustration of $ \barI_{\barhatp} $ being partitioned into 
$ \barI_{\barhatp}^{1} $, $  \barI_{\barhatp}^{2} $ and $  \barI_{\barhatp}^{3} $ using $ \barI_{\barp} $.}
\label{fig:bin1}
\end{figure}
\begin{itemize}
\item $ \barI^1_{\barhatp} $, composed of intervals 
that are equal to some interval in $ \barI_{\barp} $,
\item $ \barI^2_{\barhatp} $, that consists of intervals 
that lie strictly within some interval in $ \barI_{\barhatpp} $,
\item $ \barI^3_{\barhatp} $, containing intervals
that are unions of more than one interval from $ \barI_{\barp} $.
\end{itemize}
This is shown in Figure~\ref{fig:bin1}.
Lemmas~\ref{lem:start},~\ref{lem:middle},~\ref{lem:end} 
address each of these intervals separately. Combining the
lemmas, 
\begin{align*}
\lone{\hatf_{\barI_{\barhatp}}-f}
&=\loneint{\barI^1_{\barhatp}}{\hatf_{\barI_{\barhatp}}-f}
+\loneint{\barI^2_{\barhatp}}{\hatf_{\barI_{\barhatp}}-f} 
+\loneint{\barI^3_{\barhatp}}{\hatf_{\barI_{\barhatp}}-f} \\
&\overset{}{\le} (r_d+1)\cdot\loneint{\barI^1_{\barhatp}}
{\starf_{\barI_{\barp}}-f}
+ \sum_{I \in \barI^{1}_{\barp}}
r_d\cdot \epsilon \sqrt{(d+1)p_I}\\
&\phantom{\le} +\frac{(r_d+1)\cdot \alpha}{\alpha-2}\cdot 
\loneint{\barI^2_{\barhatp}}
{\starf_{\barI_{\barp}}-f}+\frac{1}{\alpha-1}
\sum_{ I\in \barI^2_{\barp} }
r_d\cdot \epsilon \sqrt{(d+1)p_I}\\
&\phantom{\le} +(r_d+1)\cdot 
\loneint{\barI^{3}_{\barhatp}}{\starf_{\barI_{\barp}}-f}
+\frac{\alpha\sqrt2+\sqrt{2}-1}{\sqrt{2}-1} \sum_{I\in \barI^{3}_{\barp}}
r_d\cdot \epsilon \sqrt{(d+1)p_I}\\
&\overset{(a)}{\le} \frac{(r_d+1)\cdot \alpha}{\alpha-2}
\lone{\starf_{\barI_{\barp}}-f}
+\frac{\alpha\sqrt2+\sqrt{2}-1}{\sqrt{2}-1}
\sum_{I\in \barI_{\barp}} 
r_d\cdot \epsilon \sqrt{(d+1)p_I},
\end{align*}
where $ (a) $ follows since $ \alpha>2 \Rightarrow
1/(\alpha-1)<1<(\alpha\sqrt2+\sqrt{2}-1)/(\sqrt{2}-1)$.
\end{Proof}

\begin{Lemma}
\label{lem:start}
For the final partition $ \barI_{\barhatp} $ in the run of $ \STITCH $
and any $ \barp\in \Delta_{\mathrm{bin,n}} $, let 
$\barI^{1}_{\barhatp}\subseteq \barI_{\barhatp} $ 
be the intervals that intersect with $ \barI_{\barp} $.
Let $\barI^{1}_{\barp}=\barI^{1}_{\barhatp}\subseteq \barI_{\barp} $ 
denote the corresponding collection in $ \barI_{\barp}$. Then,
\[\loneint{\barI^1_{\barhatp}}{\hatf_{\barI_{\barhatp}}-f}
\le (r_d+1)\cdot\loneint{\barI^1_{\barhatp}}
{\starf_{\barI_{\barp}}-f}
+ \sum_{I \in \barI^{1}_{\barp}}
r_d\cdot \epsilon \sqrt{(d+1)p_I},
\]
\end{Lemma}
\begin{Proof}
Follows from Theorem~\ref{thm:polyblack}
and noticing that intervals in $ \barI^{1}_{\barhatp} $ 
and $ \barI^{1}_{\barhatpp} $ coincide.
\end{Proof}

\begin{Lemma}
\label{lem:middle}
For the final partition $ \barI_{\barhatp} $ 
in the run of $ \STITCH $ and any 
$ \barp\in \Delta_{\mathrm{bin,n}} $, let 
$\barI^{2}_{\barhatp}\subseteq \barI_{\barhatp} $ 
be the intervals that do not intersect with, and strictly lie in 
some interval in $ \barI_{\barp} $.
Let $\barI^{2}_{\barp}\subseteq \barI_{\barp} $ 
be the corresponding intervals that 
contain $ \barI^{2}_{\barhatp}$. Then,
\begin{align*}
\loneint{\barI^2_{\barhatp}}{\hatf_{\barI_{\barhatp}}-f}
&\overset{}{\le} \frac{(r_d+1)\cdot \alpha}{\alpha-2}\cdot
\loneint{\barI^2_{\barhatp}}
{\starf_{\barI_{\barp}}-f}+\frac{1}{\alpha-1}
\sum_{ I\in \barI^2_{\barp} }
r_d\cdot \epsilon \sqrt{(d+1)p_I}.
\end{align*}
\end{Lemma}
\begin{Proof}
Notice that all intervals in $ \barI^2_{\barhatp} $ are strictly 
contained within some interval in $ \barI^{2}_{\barp} $. 
Using this, we further partition 
$ \barI^2_{\barhatp} $ using intervals in $ \barI^2_{\barp} $. 
Fix an $ I\in \barI^{2}_{\barp} $ and let
$ \barI \in \barI^2_{\barhatp} $ be intervals
whose union gives $ I$. 
Let $ \barhatp_{\barI}\subseteq\barhatp $ denote the 
empirical probabilities corresponding to $ \barI $
and let $ p_I $ denote the empirical probability 
under $ I $.
\begin{figure}[H]
\centering
\includegraphics[scale=0.85]{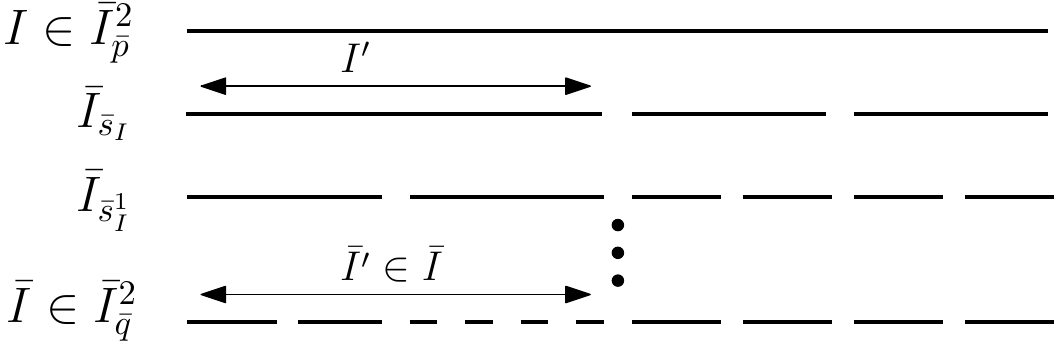}
\caption{Illustration of $ \barI\in \barI^{2}_{\barhatp}$, $ \barI_{\barhats_{I}} $ and
$ \barI_{\barhats_{I}^{1}} $ corresponding to a
particular $ I\in \barI^{2}_{\barp}$.}
\label{fig:bin2}
\end{figure}
While $ \barhatp_{\barI} $ is a sub-distribution
in general, WLOG assume $ \barhatp_{\barI} $ is 
a distribution.
Now, at some point in the run of $\STITCH$,
$ \MERGE $ was called with $ \hatf_{I,\EQ} $,
$ \barI $, 
$ \barhatp_{\barI}  $, and it was in-turn declared 
that $ \barI $ was not to be merged into $ I $. 
Therefore, for the 
$ \lemerge_{\barI,\gamma }(\hatf_{I,\EQ}) $
attaining binary distribution,
$ \barhats_{I} \in \Delta_{\mathrm{bin,n}, \ge \barhatp_{\barI}  }$,
$ \del_{I_{\barhats_{I}}}(\hatf_{I,\EQ}) - \thr_{\barhats_{I},\gamma}\ge 0$.
It follows that 
\begin{align*}
\sum_{\hats \in \barhats_{I}}
\alpha \cdot r_d\cdot \epsilon \sqrt{(d+1)\hats}
&=\thr_{\barhats_{I},\gamma}\le \del_{I_{\barhats_{I}}}(\hatf_{I,\EQ}) \\
&=\loneint{I}
{\hatf_{I,\EQ}-\hatf_{\barI_{\barhats_{I}}}}\\
&\le \loneint{I}{\hatf_{I,\EQ}-f}
+\loneint{I}{f-\hatf_{\barI_{\barhats_{I}}}}\\
&\overset{(a)}{\le}
(r_d+1)\cdot \loneint{I}{\starf_{I}-f}
+r_d\cdot \epsilon \sqrt{(d+1)p_I}\\
&\phantom{\le}+(r_d+1)\cdot
\loneint{I} {\starf_{\barI_{\barhats_{I}}}-f}
+\sum_{\hats \in {\barhats_{I}}} 
r_d\cdot \epsilon \sqrt{(d+1)\hats}\\
&\overset{(b)}{\le} 2(r_d+1)\cdot \loneint{I}{\starf_{I}-f}
+r_d\cdot \epsilon \sqrt{(d+1)p_I}+\sum_{\hats \in {\barhats_{I}}} 
r_d\cdot \epsilon \sqrt{(d+1)\hats},
\end{align*}
where $ (a) $ follows from in Theorem~\ref{thm:polyblack}, 
$ (b) $ follows since $ I $ being the union of 
$ \barI_{\barhatp_{I}} $ is also the union of 
$ \barI_{\barhats_{I}} $,
and $\starf_{I} $ is therefore a coarser approximation 
to $ f $ than $ \starf_{\barI_{\barhats_{I}}} $, 
giving rise to a larger $ \ell_{1} $ distance.
Rearrange this to obtain
\begin{equation}
\label{eqn:middleone}
\sum_{\hats \in \barhats_{I}}
r_d\cdot \epsilon \sqrt{(d+1)\hats}
\le 
\frac{1}{{\alpha-1} }\cdot\left(
2(r_d+1)\cdot \loneint{I}{\starf_{I}-f}
+r_d\cdot \epsilon \sqrt{(d+1)p_I}\right).
\end{equation}
Consider a fixed $ I' \in \barI_{\barhats_{I}}$ and let 
$ \barI' \in \barI$ be the intervals 
under $ \barI $ whose union gives $ I' $.
%
We recursively use the same argument to 
bound the LHS of Equation~\eqref{eqn:middleone}.
This is shown for the leftmost interval of $ \barI_{\barhats_{I}} $
in Figure~\ref{fig:bin2}.
Let $ \barhatp_{\barI'} $ be the corresponding 
probabilities under $ \barI'  $ and let $ \hats_{I'} $
denote the empirical probability under $ I' $.
Notice that in some previous step of $ \STITCH $, 
as was for $ I $, $ \MERGE $ was invoked with 
$ \hatf_{I',\EQ} $, $ \barI' $, $ \barhatp_{\barI'} $, 
for which 
$ \lemerge_{\barI',\gamma}(\hatf_{I',\EQ})\ge 0 $. 
Repeat the same procedure as above to obtain
\begin{align}
\nonumber
\sum_{\hats \in \barhats_{I'}}
r_d\cdot \epsilon \sqrt{(d+1)\hats}
&\le
\frac{1}{{\alpha-1} }\cdot\left(
2(r_d+1)\cdot \loneint{I'}{\starf_{I'}-f}
+r_d\cdot \epsilon \sqrt{(d+1)\hats_{I'}}\right)\\
\label{eqn:middleoneandhalf}
&\overset{(a)}{\le} \frac{1}{{\alpha-1} }\cdot\left(
2(r_d+1)\cdot \loneint{I'}{\starf_{I}-f}
+r_d\cdot \epsilon \sqrt{(d+1)\hats_{I'}}\right),
\end{align}
where $ \barhats_{I'} $ here is the binary distribution
which attains 
$ \lemerge_{\barI',\gamma}(\hatf_{I',\EQ}) $,
and $ (a) $ 
follows because $ I' $ being an interval within $ I $,
$ \starf_I $ is a coarser approximation to $ f $
than $ \starf_{I'} $. 
Summing Equation~\eqref{eqn:middleoneandhalf} 
for each such $ I' $, accumulate the distribution 
$ \barhats^{1}_{I} \ed (\cup_{I'\in \barI_{\barhats_{I}}}\barhats_{I'}) $,
and using Equation~\eqref{eqn:middleone}, the inequality,
\begin{equation}
\label{eqn:middletwo}
\sum_{\hats\in \barhats^{1}_{I}}
r_d\cdot \epsilon \sqrt{(d+1)\hats} 
\le \left(\frac{1}{\alpha-1} + \frac{1}{(\alpha-1)^2}\right)\cdot
2(r_d+1)\cdot \loneint{I}{\starf_{I}-f}
+\frac{1}{\alpha-1} \cdot r_d\cdot \epsilon \sqrt{(d+1)p_I}.
\end{equation}
Notice that while both 
$ \barhats_{I},\barhats^{1}_{I}\in \Delta_{\mathrm{bin,n},\ge \barhatp_I} $,
$\barhats^{1}_{I}$ is at least one notch closer to $ \barhatp_{I} $ as
$ \barhats^{1}_{I}\in \Delta_{\mathrm{bin,n}, < \barhats_{I}} $.
Since the number of binary distributions is finite,
on recursively using this argument,
summation across $ \barhatp_{I} $ is eventually
obtained on the LHS. Iterating on this procedure yields 
the upper bound
\begin{align*}
\sum_{\hatp \in \barhatp_{\barI}}
r_d\cdot \epsilon \sqrt{(d+1)\hatp}
&\le \left(\frac{1}{\alpha-1} + \frac{1}{(\alpha-1)^2}+\cdots\right)\cdot
2(r_d+1)\cdot \loneint{I}{\starf_{I}-f}
\\&
\phantom{\le}+\frac{1}{\alpha-1} r_d\cdot \epsilon \sqrt{(d+1)p_I}\\
&\overset{(a)}{\le} \frac{2(r_d+1)}{\alpha-2}\cdot
\loneint{I}{\starf_{I}-f}+\frac{1}{\alpha-1}\cdot
r_d\cdot \epsilon \sqrt{(d+1)p_I},
\end{align*}
where $ (a) $ follows since $ \alpha>2 $.
Repeating this argument across each 
$I \in \barI^2_{\barp} $,
\begin{equation}
\label{eqn:middlethree}
\sum_{I\in \barI^2_{\barhatp}}
r_d\cdot \epsilon \sqrt{(d+1)\hatp_I}
\le \frac{2(r_d+1)}{\alpha-2}\cdot
\loneint{\barI^2_{\barhatp}}
{\starf_{\barI_{\barp}}-f}+\frac{1}{\alpha-1}
\sum_{ I\in \barI^2_{\barp} }
r_d\cdot \epsilon \sqrt{(d+1)p_I}.
\end{equation}
This finally gives us
\begin{align*}
\loneint{\barI^2_{\barhatp}}{\hatf_{\barI_{\barhatp}}-f}
&\overset{(a)}{\le} (r_d+1)\cdot
\loneint{\barI^2_{\barhatp}}{\starf_{\barI_{\barhatp}}-f}
+\sum_{I \in \barI^{2}_{\barhatp}}
r_d\cdot \epsilon \sqrt{(d+1)\hatp_I}\\
&\overset{(b)}{\le} (r_d+1)\cdot
\loneint{\barI^2_{\barhatp}}{\starf_{\barI_{\barp}}-f}
+\sum_{I \in \barI^{2}_{\barhatp}}
r_d\cdot \epsilon \sqrt{(d+1)\hatp_I}\\
&\overset{}{\le} (r_d+1) \left(1+\frac{2}{\alpha-2}\right)
\loneint{\barI^2_{\barhatp}}
{\starf_{\barI_{\barp}}-f}+\frac{1}{\alpha-1}
\sum_{ I\in \barI^2_{\barp} }
r_d\cdot \epsilon \sqrt{(d+1)p_I},
\end{align*}
where $ (a) $ follows from Theorem~\ref{thm:polyblack}, 
$ (b) $ follows since, by definition, intervals in 
$ \barI^{2}_{\barhatp} $ lie within those in 
$ \barI^{2}_{\barp} $, and thus $ \starf_{\barI_{\barp}} $
is a coarser approximation to $ f $ than $ \starf_{\barI_{\barhatp}} $
in $ \barI_{\barhatp}^{2} $,
and finally $ (c) $ follows by plugging in Equation~\eqref{eqn:middlethree}.
\end{Proof}

\begin{Lemma}
\label{lem:end}
For the final partition $ \barI_{\barhatp} $ in the run of $ \STITCH $
and any $ \barp\in \Delta_{\mathrm{bin,n}} $, let 
$\barI^{3}_{\barhatp}\subseteq \barI_{\barhatp} $ 
be intervals that are unions of more 
than one interval from $ \barI_{\barp} $.
Let $\barI^{3}_{\barp}\subseteq \barI_{\barp} $
be the corresponding intervals whose union gives 
$ \barI^{3}_{\barhatp}$. Then,
\begin{align*}
\loneint{\barI^{3}_{\barhatp}}{\hatf_{\barI_{\barhatp}}-f}
\le (r_d+1)\cdot \loneint{\barI^{3}_{\barhatp}}{\starf_{\barI_{\barp}}-f}
+\frac{\alpha\sqrt2+\sqrt{2}-1}{\sqrt{2}-1} \sum_{I\in \barI^{3}_{\barp}}
r_d\cdot \epsilon \sqrt{(d+1)p_I}.
\end{align*}
\end{Lemma}
\begin{Proof}
Fix an $ I \in \barI^{1}_{\barhatp}$
and let $ \hatp_{I} $ be its empirical probability. 
Let $ \barI_{\barp, I} \in \barI_{\barp} $ 
indicate intervals
under $ \barI_{\barp} $ whose union gives 
$ I $ and let $ \barp_{I} \subseteq \barp$ denote
the corresponding empirical probabilities under
$ \barI_{\barp,I}$.
In run of $ \STITCH $, let the interval 
collection that was merged to create $ I $ be denoted 
by $ \barI_{\barhatp,I} $, and its collection of empirical probabilities 
by $ \barhatp_{I}$. While $ \barp_{I} $ is a 
sub-distribution in general, WLOG assume
it is a distribution. This also implies 
$ \barhatp_{I}$ is a distribution.

Using $ \barI_{\barhatp,I}$, separate $ \barI_{\barp,I} $ into
\begin{itemize}
\item $ \barI_{\barp,I}^{1} $, consisting of intervals
in $ \barI_{\barp,I}$ that are equal to, or unions of 
intervals from $ \barI_{\barhatp,I} $.
\item $ \barI_{\barp,I}^{2} $, intervals in $ \barI_{\barp,I}$
that lie strictly inside some interval in $ \barI_{\barhatp,I} $.
\end{itemize}

Let $ \barI_{\barhatp,I}^2\subseteq $ 
be the corresponding intervals in 
$ \barI_{\barhatp,I} $ 
that contain $ \barI_{\barp,I}^{2} $.
Let $ \barp^{1}_{I} $, $ \barp^{2}_I $ be empirical probabilities
corresponding to $ \barI_{\barp,I}^{1} $, $ \barI_{\barp,I}^{2} $
respectively. 
Similarly let $ \barhatp^{2}_I $ correspond to $ \barI_{\barhatp,I}^{2} $.
This is shown in Figure~\ref{fig:bin3}, where the arrow indicates
the collection of intervals merged by $ \STITCH $.
\begin{figure}[H]
\centering
\includegraphics[scale=0.85]{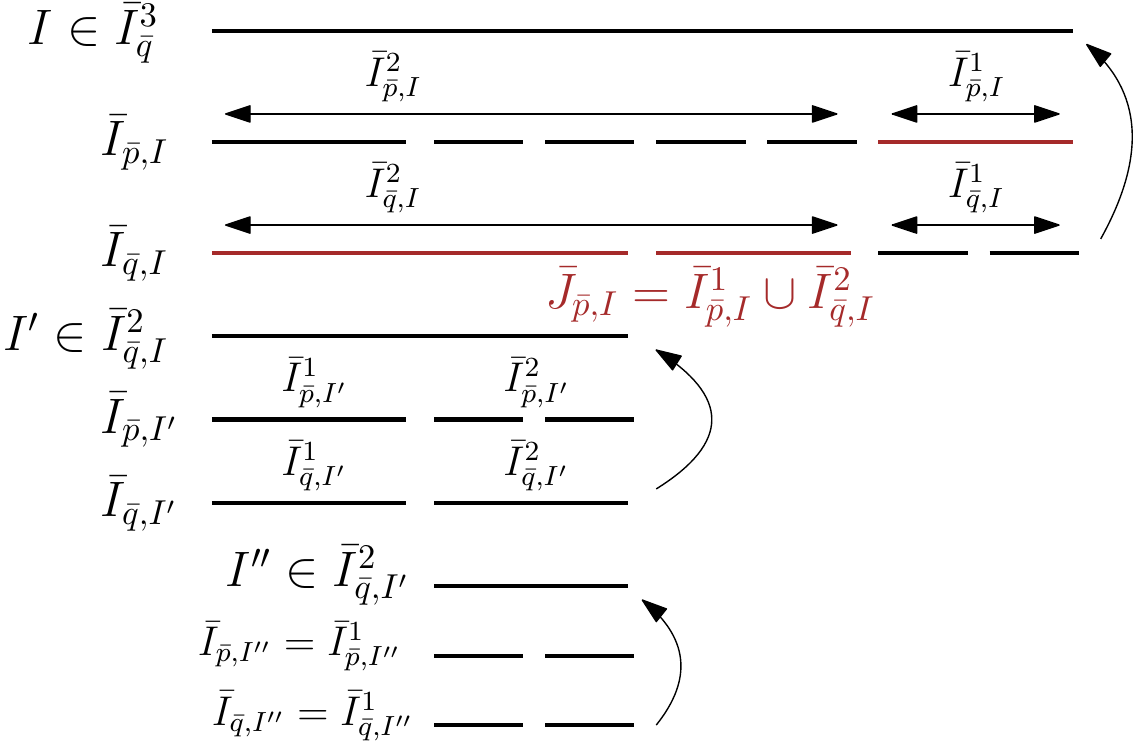}
\caption{Illustration of proof construction for a particular $ I\in \barI^{3}_{\barhatp}$.}
\label{fig:bin3}
\end{figure}
Modify $ \barI_{\barp,I} $ to obtain 
a new partition 
$ \barJ_{\barp,I} \ed \barI_{\barp,I}^{1} \cup \barI_{\barhatp,I}^2$.
Now each interval in $ \barJ_{\barp,I} $ is equal to, 
or is a union of intervals from $ \barI_{\barhatp,I} $. 
Equivalently, if $ \barhats $ is the 
empirical distribution corresponding to 
$ \barJ_{\barp,I} $, 
$ \barhats \in \Delta_{\mathrm{bin,n},\ge \barhatp_{I}} $.
Since $ I $ was merged when the merging routine
was called with $ \hatf_{I,\EQ} $,
$ \barI_{\barhatp,I}$, $ \barhatp_{I}$,
it implies
$ \thr_{\barhats,\gamma} \ge \del_{I_{\barhats}}(\hatf_{I,\EQ}) $. 
Therefore
\begin{align}
\nonumber
\loneint{I}{\hatf_{\barI_{\barhatp}}-f}&\le \loneint{I}{\hatf_{\barJ_{\barp,I}}-f} 
+ \loneint{I}{\hatf_{\barI_{\barhatp}}-\hatf_{\barJ_{\barp,I}}} \\
\nonumber
&\overset{(a)}{=} \loneint{I}{\hatf_{\barJ_{\barp,I}}-f}  + \del_{I_{\barhats}}(\hatf_{I,\EQ}) \\
\nonumber
&\le \loneint{I}{\hatf_{\barJ_{\barp,I}}-f}  + \thr_{{\barhats},\gamma}\\
\nonumber
&\overset{(b)}{=} \loneint{\barJ_{\barp,I}}{\hatf_{\barJ_{\barp,I}}-f}  
+\thr_{{\barhats},\gamma}\\
\nonumber
&= \loneint{\barJ_{\barp,I}}{\hatf_{\barJ_{\barp,I}}-f}  + \alpha
\sum_{s\in\barhats}
r_d\cdot \epsilon \sqrt{(d+1)s}\\
\nonumber
&\overset{(c)}{=} \loneint{\barI^1_{\barp,I}}{\hatf_{\barI^{1}_{\barp,I}}-f} 
+ \alpha\sum_{p\in\barp^{1}_{I}}
r_d\cdot \epsilon \sqrt{(d+1)p}\\
\nonumber
&\phantom{=}
+ \loneint{\barI^2_{\barhatp,I}}{\hatf_{\barI^{2}_{\barhatp,I}}-f}
+ \alpha\sum_{\hatp\in\barhatp^{2}_{I}}
r_d\cdot \epsilon \sqrt{(d+1)\hatp}\\
\nonumber
&\overset{(d)}{=} \loneint{\barI^1_{\barp,I}}{\hatf_{\barI_{\barp}}-f}  
+ \alpha\sum_{p\in\barp^{1}_{I}}
r_d\cdot \epsilon \sqrt{(d+1)p}\\
\label{eqn:threeone}
&\phantom{=}
+ \loneint{\barI^2_{\barhatp,I}}{\hatf_{\barI^{2}_{\barhatp,I}}-f}
+ \alpha\sum_{\hatp\in\barhatp^{2}_{I}}
r_d\cdot\epsilon \sqrt{(d+1)\hatp},
\end{align}
where $ (a) $ follows since by definition,
$\hatf_{\barI_{\barhatp}} = \hatf_{I,\EQ} $
in interval $ I $,
$ (b) $ follows since $ \barJ_{\barp,I} $ being 
a partition of $ I $ lies in the same region as $ I $, 
$ (c) $ follows since
$ \barJ_{\barp,I} = \barI_{\barp,I}^{1} \cup \barI_{\barhatp,I}^2$,
and $ (d) $ follows since 
$ \hatf_{\barI_{\barp}} = \hatf_{\barI^{1}_{\barp,I}}$
in $ \barI_{\barp,I}^{1} $ as $ \barI^{1}_{\barp,I}\subseteq \barI_{\barp}$. 

Now consider an interval $ I' \in \barI_{\barhatp,I}^{2} $. 
Since $\barI_{\barhatp,I}^{2}\subseteq \barI_{\barhatp,I}$, 
and since $ \barI_{\barhatp,I} $, by definition, are
intervals that were merged to produce $ I $,
it follows that $ I' $ in turn was an interval that was merged into in
some previous step of $ \STITCH $. 
As before, let the intervals that were merged
to generate $ I' $ be denoted by $ \barI_{\barhatp,I'} $.
Further, by definition of
$ \barI_{\barhatp,I}^{2}$, all intervals in it
occur as unions of those in $ \barI^{2}_{\barp,I}$, 
and so does $ I' $.
Let $ \barI_{\barp,I'} \subseteq \barI^{2}_{\barp,I} $ be 
these intervals whose union gives $ I' $.
Repeat the same argument as above to obtain
\begin{align}
\nonumber
\loneint{I'}{\hatf_{\barI^{2}_{\barhatp}}-f}
&\le  \loneint{\barI^1_{\barp,I'}}{\hatf_{\barI_{\barp}}-f}  
+ \alpha\sum_{p\in\barp^{1}_{I'}}
r_d\cdot \epsilon \sqrt{(d+1)p}\\
\label{eqn:threetwo}
&\phantom{\le}
+ \loneint{\barI^2_{\barhatp,I'}}{\hatf_{\barI^{2}_{\barhatp,I'}}-f}
+ \alpha\sum_{\hatp\in\barhatp^{2}_{I'}}
r_d\cdot \epsilon \sqrt{(d+1)\hatp},
\end{align}
where each of  $ \barI^{1}_{\barp,I'} $, $ \barp^{1}_{I'} $, 
$ \barI^2_{\barhatp,I'} $ and $ \barhatp^{2}_{I'}$ are 
defined in exactly the same manner as was for $ I $,
but by replacing $ I' $ in all definitions.
Since $ \barI^{1}_{\barp,I'} 
\subseteq \barI^{2}_{\barp,I}\subseteq \barI_{\barp}$,
substituting Equation~\eqref{eqn:threetwo}
into~\eqref{eqn:threeone},
a larger portion of $ I $ is bounded using the 
difference $ \loneint{}{\hatf_{\barI_{\barp}} -f}$.
Upon repeating the same argument for all
$ \loneint{}{\hatf_{\barI_{\barhatp}}-f}$ terms that 
remain, a bound on the RHS is obtained that 
consists exclusively of $ \loneint{}{\hatf_{\barI_{\barp}}-f} $.
The entire procedure is shown in Figure~\ref{fig:bin3}.

Further, from Lemma~\ref{lem:treegeosum}, 
the sum of all the $ \epsilon $-deviation terms 
that results on the RHS from repeating the 
argument is bounded by $ \sqrt2/(\sqrt{2}-1) $ times
the total $ \epsilon $-deviation in $ \barI_{\barp}^{3} $. This results in
\begin{align*}
\loneint{I}{\hatf_I-f} &\le \loneint{I}{\hatf_{\barI_{\barp}}-f}
+ \frac{\alpha}{\sqrt{2}-1} \sum_{p\in\barp_{I}}
r_d\cdot \epsilon \sqrt{(d+1)p}\\
&\overset{(a)}{\le} (r_d+1)\cdot \loneint{I}{\starf_{\barI_{\barp}}-f}
+ \left(1+\frac{\alpha\sqrt2}{\sqrt{2}-1}\right) \sum_{p\in\barp_{I}}
r_d\cdot\epsilon \sqrt{(d+1)p},
\end{align*}
where $ (a) $ follows from Theorem~\ref{thm:polyblack}.
Repeating across $I\in \barI^{3}_{\barhatp}  $ gives 
\begin{align*}
\loneint{\barI^{3}_{\barhatp}}{\hatf_{\barI_{\barhatp}}-f}
\le (r_d+1)\cdot \loneint{\barI^{3}_{\barhatp}}{\starf_{\barI_{\barp}}-f}
+ \left(1+\frac{\alpha\sqrt2}{\sqrt{2}-1}\right) \sum_{I\in \barI^{3}_{\barp}}
r_d\cdot \epsilon \sqrt{(d+1)p_I}.
\end{align*}
\end{Proof}

\begin{Lemma}
\label{lem:treegeosum}
Suppose in the run of $ \STITCH $, a collection of consecutive 
intervals $ \barI_1 $ was merged in $ k-1 $ steps 
to generate $ \barI_k $, and suppose 
$ \barI_2\upto \barI_{k-1} $ are 
the intermediate interval collections. Then, 
\[\sum_{i=1}^k
\sum_{I\in \barI_i}\sqrt{\hatp_I}
\le \sum_{I\in \barI_1} \frac{\sqrt2}{\sqrt{2}-1}\sqrt{\hatp_I}.
\]
\end{Lemma}
\begin{Proof}
WLOG assume $ \barhatp_{\barI_{1}} $ is a distribution, 
which also implies $ \barhatp_{\barI_{i}} $ is a distribution 
$ \forall i\in \{2\upto k\} $. Notice that
for any $ i\in \{2\upto k\} $,
$ \barhatp_{\barI_{i}}\in \Delta_{\mathrm{\mathrm{bin,n}},
\ge \barhatp_{\barI_{i-1}}} $.
Thus $ |\barI_{i}| \le 1/2\cdot  |\barI_{i-1}|$, 
where $ |\barI| $ denotes the number of intervals in $ \barI $.
By concavity of $ \sqrt{x} $ for $ x\ge 0 $, the sum 
$ \sum_{I\in \barI_{i}}\sqrt{\hatp_I} $ is maximized
for a given $ \barhatp_{\barI_{i-1}} $, 
if $  |\barI_{i-1}| = 2 \cdot |\barI_{i}|$.
Since this equality is attained iff $ \barhatp_{\barI_{i-1}} $ is
the uniform distribution over $ |\barI_{i-1}| $ elements
and $ \barhatp_{\barI_{i}} $ is uniform over 
$ |\barI_{i}| = 1/2\cdot |\barI_{i-1}| $ elements,
\[\sum_{I\in \barI_i}\sqrt{\hatp_I} 
\le \frac{1}{\sqrt{2}}\sum_{I\in \barI_{i-1}}\sqrt{\hatp_I}.\]
This implies 
\[\sum_{i=1}^k
\sum_{I\in \barI_i}\sqrt{\hatp_I}
\le \sum_{i=1}^k \left(\frac{1}{\sqrt{2}}\right)^{i-1}\cdot
\sum_{I\in \barI_1} \sqrt{\hatp_I}
\le \frac{\sqrt2}{\sqrt{2}-1}\sum_{I\in \barI_1} \sqrt{\hatp_I}.
\]
\end{Proof}\label{appen:stitch}
\section{Additional Lemmas}
\begin{Lemma}
\label{lem:boss}
Adapting~\cite{bousquet19} to achieve a factor-$2$ approximation 
for $\cP_2$ results in $\epsilon_n=\tilde{\cO}(n^{-1/4})$.
\end{Lemma}
\begin{Proof}
WLOG fix the interval be $ [0,1] $. Further if
$ \poly\in \cP_{2} $ has an $ \ell_{1} $ norm $> 2 $,
it follows that for any distribution $f $, 
$ \lone{\poly-f} \ge \lone{\poly}-\lone{f}\ge 1 $. Thus 
WLOG restrict the $ \cP_{2} $ to the
subset $\cQ \ed \{\poly \in \cP_{2}: 
\lone{\poly}\le 2, \poly \ge 0\}  $

Let $ \cD_{\epsilon} $ be an arbitrary $ \epsilon/2 $ 
cover of $ \cQ$. Thus for the estimate 
$ \hatf_{\mathrm{BK}} $ output 
by~\cite{bousquet19},
\begin{equation}
\label{eqn:boss}
\EE\lone{\hatf_{\mathrm{BK}}-f}
= 2\OPT_{\cD_{\epsilon}}(f) + \tilde{\cO}\left(|\cD_{\epsilon}|^{1/5}/n^{2/5}\right)
\le 2\OPT_{\cP_{2}}(f) +\epsilon+ \tilde{\cO}\left(|\cD_{\epsilon}|^{1/5}/n^{2/5}\right).
\end{equation}

For
$ \bar{c}=(c_0,c_1)\in \RR^2 $, let $ \poly_{\bar{c}}\ed c_0+c_1x+c_{2}x^2$.
Consider $ \cC_{\epsilon} $, 
a subset of $\cQ$ defined as
$ \cC_{\epsilon} 
\ed \{\poly_{\bar{c}} \in \cQ: c_i= \lambda_i \epsilon, \ 
\lambda_{i}\in \ZZ, \  i \in \{0,1,2\} \}$. It is easy to see that
$ |\cC_{\epsilon}|\ge  \Omega(1/\epsilon^3) $.
Since the $\ell_1$ norm between any two members in
$\cC_{\epsilon}$ is at least $ \epsilon/2$, 
$ |\cD_{\epsilon}|\ge |\cC_{\epsilon}| \ge  \Omega(1/\epsilon^3)$.

Optimizing Equation~\eqref{eqn:boss} w.r.t. $ \epsilon $
results in 
\[
\EE\lone{\hatf_{\mathrm{BK}}-f}
\le 2\OPT_{\cP_{1}}(f) + \tilde{\cO}(n^{-1/4}).
\]
\end{Proof}

\begin{Lemma}
\label{lem:gauss}
Let $ f $ be a Gaussian distribution.
Then for a constant $ d $,
$ \OPT_{\cP_{t,d}}(f) = \cO(1/t^{d-1})$.
\end{Lemma}
\begin{Proof}
Let $ t = t_{1}-2 $. 
WLOG assume $ f $ has mean $ 0 $ and variance $ 1 $
so that $ f = 1/\sqrt{2\pi}\cdot e^{-x^2/2} $.
Fix an $ L>0 $. 
Divide $ [-L,L] $
into $ t $ equal sized intervals of length 
$ l\ed 2L/t$.
Let $ \poly_{d} $ be the $ t $-piecewise
order $ d $ Taylor polynomial of $ f $ on that interval.
Then for any $ x\in \RR $,
$ |\poly_{d}(x)-f (x)|\le f^{d+1}(c) l^{d+1}/(d+1)!$, 
where $ 0\le c\le l $.

Since $  f^{d+1}(x) = H_{d+1}(x)e^{-x^{2}/2} $ 
where $ H_{d+1} $ is the $ d+1 $th-order Hermite polynomial,
standard bounds~\cite{ilya04} imply that there exists a constant $c_d : f^{d+1}(x) \le c_d, \forall x \in \RR$.

Observe that on the interval, $ (L,\infty)$,
from standard sub-gaussian inequalities,
$$ \int_{L}^{\infty} 
\frac{1}{\sqrt{2\pi}}e^{-x^{2}/2}dx
\le e^{-L^2/2}.
$$
Similarly for $ (-\infty,L)$, extend $ \poly_{d} $ to these 
intervals to obtain a $ t+2 $-piecewise polynomial.
Then 
$$ \lone{\poly_{d}-f} \le 
t\cdot c_d \frac{l^{d+1}}{(d+1)!} + e^{-L^2/2}
=c_d \frac{(2L)^{d+1}}{t^{d}(d+1)!} + e^{-L^2/2}$$

Choosing $ L = \cO({\sqrt{2d\log t}} )$ gives 
$  \lone{\poly_{d}-f}\le \cO\left(\left((\log t)^{\frac{d+1}{2}}+1\right)/t^{d}\right) = \cO(1/t^{(d-1)})$.
\end{Proof}\label{appen:extra}
\section{$ \STITCH $ and $\MERGE$ Algorithms}
\label{appen:algomain}
This section provides a detailed description of 
$ \STITCH $ and $ \MERGE $, the main routines 
of $ \SURF $. We also restate the necessary 
definitions.

\subsection{The $\STITCH$ Routine}
\label{appen:alg1}
$ \STITCH$ receives as input, $ X^{n-1} $ and 
parameters $ d $, $\alpha$, $ \epsilon $.
The routine operates in $i\in\{1\upto D\} $ steps.
Define $ D(i)\ed D-i$ and let
\[ \bar{u}_i \ed \left({1}/{2^{D(i)}}\upto {1}/{2^{D(i)}}\right), \
\barI_{\bar{u}_i} =(I_{\bar{u}_i,1}\upto I_{{\bar{u}_i}, 2^{D(i)}}).\]
Initialize $\barhatp_0\gets  \baru{n}$. 

Start with $ i =1  $ and assign $ \barhats \gets \barhatp_{i-1}$.
In each step, the routine maintains this
$\barhats = \barhatp_{i-1}\in \Delta_{\mathrm{bin,n},\le \bar{u}_i}$.
This can be seen from the initialization above
for $ i=1 $ since $ \bar{u}_1 = (2/n\upto 2/n) $, 
and verified for $ i>1 $. Thus, using $ \barI_{\bar{u}_{i}} $, 
we may separate
\[
\barI_{\barhats}
=(\barI_{\barhats,1}\upto \barI_{\barhats,2^{D(i)}}),\
\barhats=(\barhats_1\upto\barhats_{2^{D(i)}}),
\]
where for each $ j\in\{1\upto 2^{D(i)}\} $,
$\barI_{\barhats,j}\subseteq \barI_{\barhats}$
are intervals in $ \barI_{\barhats} $ whose union 
gives $ I_{\bar{u}_i,j}\in \barI_{\bar{u}_i} $. 
Let $\barhats_j\in \barhats$ denote the empirical probabilities in $ \barhats $
corresponding to intervals in $ \barI_{\barhats,j}$.
Notice that the sum of all probabilities in $ \barhats_j $, 
$\sum_{\hats\in \barhats_j}\hats = 1/2^{D(i)} $.
Therefore the scaled $ 2^{D(i)}\barhats_j $
is an empirical distribution.
For brevity, let the polynomial estimate
output by $ \EQ $ on $ I_{{\bar{u}_i},j} $,
be denoted by
\[ \hatf_{I_j}\ed \hatf_{I_{{\bar{u}_i},j},\EQ}. \]
Starting with $ j=1 $, invoke $ \MERGE $ with arguments, 
the polynomial estimate $ \hatf_{I_j} $,
intervals $ \barI_{\barhats,j} $ and the empirical
distribution $ 2^{D(i)}\barhats_j $, samples 
$ X^{n-1}_{i,j}
\subseteq X^{n-1} $ that lie in $ I_{\bar{s},j} $, and parameters $ d $,
\[\gamma \ed 
\alpha\cdot r_d \cdot \epsilon \sqrt{d+1}.
\]
This parameter, $\gamma$, is used to
tune the bias-variance trade-off. 
As will be shown subsequently, 
if $ \gamma \rightarrow\infty $,
$ \barI_{\barhats,j} $ will be merged, resulting in 
an estimate with a larger bias but smaller variance.
A small $ \gamma $ has the opposite effect.

If $ \MERGE(\hatf_{I_j}, \barI_{\barhats,j}, 
2^{D(i)}\barhats_j, X^{n-1}_{i,j},d,\gamma)\le 0 $,
merge $\barI_{\barhats,j}  $ 
into a single interval $ I_{\bar{u}_i,j} $. Accomplish this by
updating $\barhats_j$ to a unitary value, 
its sum, ($ 1/2^{D(i)}$).  
Otherwise, maintain $ \barhats $ as is.
Increment $j$ within the range 
$\{1 \upto 2^{D(i)}\}$ and repeat this procedure.

After the entire run in $ j $ is complete, 
update $\barhatp_i\leftarrow \barhats$.
If $ D(i)=D-i>0$, increment $i$ and 
repeat the same steps.
Otherwise, if $ D(i)=0$ or in other words if $ i=D$, 
$ \STITCH $,  
and in turn, $ \SURF $ outputs the piecewise 
estimate on $ \barI_{\barhatp_{D}} $, i.e. 
$ \hatf_{\text{SURF}}   =\hatf_{\barI_{\barhatp_{D}},\EQ } $.

At each step $ i\in \{1\upto D\} $, $ \STITCH $ calls 
$ \MERGE $ on $2^{D(i)} $ intervals,
each consisting of $ 2^i $ samples. 
Thus each step of $ \STITCH $ takes
$ \cO(2^{D(i)}\cdot (d^{\tau}+\log (2^i)) \cdot 
2^i) = \cO((d^{\tau}+\log n )2^D)$ time. 
The total time complexity is therefore 
$\cO((d^{\tau}+\log n )2^D D) = \cO((d^{\tau}+\log n)n\log n ) $.

\begin{algorithm}[tb]
   \caption{$ \STITCH $}
   \label{alg:STITCH}
\begin{algorithmic}
   \STATE {\bfseries Input:} $ X^{n-1} $, $ d $, $\alpha$, $ \epsilon $
   \STATE Initialize $D= \log n, \barhatp= \baru{n}$, $\gamma \gets\alpha\cdot  r_d \epsilon\cdot \sqrt{d+1}$
   \FOR{$i=1$ {\bfseries to} $D$}
	\STATE $ D(i)\gets D-i  ,\ \barhats\gets \barhatp$
    \FOR{$j=1$ {\bfseries to} $2^{D(i)}$}
	\IF{$ \MERGE(\hatf_{I_j},   \barI_{\barhats,j},  2^{D(i)}\barhats_j , X^{n-1}_{i,j},d, \gamma)$ 
	$\le 0$}
   \STATE $ \barhats_j \gets (1/2^{D(i)})$
   \ENDIF
	\ENDFOR
	\STATE $ \barhatp \gets \barhats $
   \ENDFOR
\STATE {\bfseries Output:} $ \barhatp $
\end{algorithmic}
\end{algorithm}
\ignore{-.5em}
\subsection{The $\MERGE$ Routine}
\ignore{-.25em}
$ \MERGE $ receives as input, a function $ \hatf $, 
an interval partition $ \barI \ed \barI_{\barhats} $ 
and the corresponding empirical 
distribution $ \barhats $, samples $ X^{m} $ that lie in $ \barI $, 
and parameters $ d $, $ \gamma $.

\ignore{Notice that the number of samples within $ I $
is given by $ \sum_{s\in\barhats}s\cdot n $.
While $ \barhats $ forms a sub-distribution in general,
WLOG assume it is a distribution, 
implying $ I $ has all $ n-1 $ samples. }

Fix a $ \barp \in \Delta_{\mathrm{bin,m},\ge \barhats} $,
and consider the piecewise polynomial estimate 
on $ \barI_{\barp} $, $ \hatf_{\barI_{\barp},\EQ} $. Define
\begin{equation}
\label{eqn:thr}
\del_{\barI_{\barp}}(\hatf)
\ed
\loneint{\barI_{\barp}}{\hatf_{\barI_{\barp},\EQ}
	-\hatf},\
\thr_{\barp, \gamma}\ed 
\sum_{p\in \barp}
\gamma \sqrt{p}.
\end{equation}
 $ \MERGE(\hatf) $ returns 
$ \lemerge_{\barI_{\barhats},\gamma}(\hatf) $, 
the largest difference between $ \del_{I_{\barp}}(\hatf) $
and $ \thr_{\barp,\gamma} $ across all
$ \barp\in \Delta_{\mathrm{bin,m}, \ge \barhats } $,
\[
\lemerge_{\barI_{\barhats},\gamma}(\hatf)
\ed
\max_{\barp \in \Delta_{\mathrm{bin,m}, \ge \barhats }}
\del_{\barI_{\barp}}(\hatf)-\thr_{\barp, \gamma}.
\]
The quantity, $ \del_{\barI_{\barp}}(\hatf) $ acts as a proxy for the increment 
in bias that results if the piecewise estimate
$ \hatf_{\barI_{\barp},\EQ} $ is merged into $ \hatf $, while 
$ \thr_{\barp, \gamma} $ accounts for the deviation in 
$\hatf_{\barI_{\barp},\EQ} $ under $ \cQ_{\epsilon} $.
Notice that for any 
$\barp \in \Delta_{\textrm{bin,m}, \ge \barhats} $,
$ \thr_{\barp, \gamma} \le \thr_{\barhats, \gamma}$.
Thus $ \lemerge_{\barI_{\barhats},\gamma}(\hatf)
\le 0 $ 
if the decrease in deviation under $ \barI = \barI_{\barhats} $ is larger 
than the increased bias under any candidate $ \barI_{\barp} $. 
This in turn signals $ \STITCH $ to merge $ \barI $.

It may be shown that if $ \barhats = \baru{m} $,
the cardinality, $ |\Delta_{\mathrm{bin,m}, \ge \barhats }|  
= \Omega(m ^c)$ for any $ c>0 $. Therefore,
naively evaluating $ \del_{I_{\barp}}(\hatf)-\thr_{\barp,\gamma} $ over each
$\barp \in \Delta_{\mathrm{bin,m}, \ge \barhats } $ 
incurs a worst case time complexity that is 
super-linear in $ m $. Instead, $ \MERGE $ uses
a simple divide-and-conquer procedure that computes 
$ \lemerge_{\barI_{\barhats},\gamma}(\hatf) $
in time $ \cO((d^{\tau}+\log m)m)$.

To describe this, notice that
if $ \barI_{\barhats} $ is a singleton $ (I) $, then $ \barhats = (1)$, implying
$\Delta_{\mathrm{bin,m}, \ge \barhats }= \{(1)\}$. 
In this case, obtain $ \hatf_{I, \EQ} \in \cP_d$
and return 
\[\lemerge_{\barI_{\barhats},\gamma}(\hatf)=
\del_{\barI_{(1)}}(\hatf)-\thr_{(1),\gamma}=
\loneint{\barI_{(1)}}{\hatf_{I, \EQ} - \hatf}-\gamma\sqrt{1}.
\]

If $ \barI_{\barhats} $ is non singleton or 
$ \barhats \neq (1) $, any 
$ \barp \in \Delta_{\mathrm{bin,m}, \ge \barhats } \setminus\{ (1)\}$ 
may be split into two sub-distributions, $ \barp_1 $, 
$ \barp_{2} $ that each sum to $1/2$. For example,
if the particular $\barp = (1/4,1/4,1/8,1/8,1/4)$,
it may be split into $\barp_1 = (1/4,1/4)$ and 
$ \barp_2 = (1/8,1/8,1/4) $. The corresponding
interval partition is also split into 
$\barI_{\barp} =  (\barI_{\barp_{1}},\barI_{\barp_{2}})$.
Since $ \barhats \neq (1)$, 
this may also be similarly split 
into $ \barhats_1 $ and $\barhats_2$. 
As a consequence, $ \barI_{\barhats}$ is also cleaved
into $ (\barI_{\barhats_1}, \barI_{\barhats_2}) $ 
corresponding to $ \barhats_1 $ and $\barhats_2$. 
Using this observation,
\begin{align*}
\max_{\barp \in \Delta_{\mathrm{bin,m}, \ge \barhats } , \ 
\barp \neq (1)}
\del_{\barI_{\barp}}(\hatf)-\thr_{\barp,\gamma}
&=\max_{\barp \in \Delta_{\mathrm{bin,m}, \ge \barhats } , \ 
\barp \neq (1)}
\del_{\barI_{\barp_1}}(\hatf)-\thr_{\barp_1,\gamma}+
\del_{\barI_{\barp_2}}(\hatf)-\thr_{\barp_2,\gamma}\\
&=\max_{\barp_1 \in \Delta_{\mathrm{bin,m}/2, \ge 2\barhats_1 } }
\del_{\barI_{\barp_1}}(\hatf)-\thr_{\barp_1,\gamma/\sqrt2}\\
&\phantom{=}+\max_{\barp_2 \in \Delta_{\mathrm{bin,m}/2, \ge 2\barhats_2 }} 
\del_{I_{\barp_2}}(\hatf)-\thr_{\barp_2,\gamma/\sqrt2}\\
&=\lemerge_{\barI_{2\barhats_1},\gamma/\sqrt2}(\hatf)
+\lemerge_{\barI_{2\barhats_2},\gamma/\sqrt2}(\hatf),
\end{align*}
where $ 2\barhats_1 $, $ 2\barhats_2 $ are the 
normalized variants of $ \barhats_1 $, $ \barhats_2 $, 
and $ \gamma $ is scaled by $ 1/\sqrt{2} $ to accommodate 
for this scaling. By evaluating
$\lemerge_{\barI_{2\barhats_1},\gamma/\sqrt2}
(\hatf) $,
$\lemerge_{\barI_{2\barhats_1},\gamma/\sqrt2}
(\hatf)$ separately, and then comparing their sum with
$ \del_{\barI_{(1)}}(\hatf)-\thr_{ (1),\gamma} $,
we allow for a recursive computation of 
$\lemerge_{\barI_{\barhats},\gamma}(\hatf) $.

Let $ X_{1}^{m} $ and $ X_2^{m}$ denote
the samples in $ \barI_{\barhats_1} $ and $ \barI_{\barhats_2} $ respectively.
Using these arguments, call $ \MERGE $ on 
$ \barI_{\barhats_1}, \barhats_{1} $ and 
$ \barI_{\barhats_2}, \barhats_{2} $,
return the maximum as shown in Algorithm~\ref{alg:COMP}.

Now $ \del_{\barI_{(1)}}(\hatf)-\thr_{ (1),\gamma} $
is calculated by obtaining $ \hatf_{I, \EQ} \in \cP_d$ 
from $ \EQ $. Since $ I $ has $ m $ samples, from 
Theorem~\ref{thm:polyblack}, this takes
$ \cO(m+d^{\tau}) $ time. Further, notice that since 
both $ \barhats_1 $ and $ \barhats_2$
sum to $ 1/2 $, the split $ \barI_{\barhats}=
(\barI_{\barhats_1},\barI_{\barhats_2}) $ occurs along
the median of $ X^{m} $. Thus $ \barI_{\barhats_1} $
and $ \barI_{\barhats_2} $ has at most half the
number of samples, $ m/2 $, and the time
complexity of $ \MERGE $, $ T(m) $, is captured by
\[T(m)\le 2T(m/2)+\cO(m+d^{\tau}),\]
implying $ T(m) = \cO((d^{\tau}+\log m)m) $.

\begin{algorithm}[tb]
   \caption{$ \MERGE $}
   \label{alg:COMP}
\begin{algorithmic}
   \STATE {\bfseries Input:} 
$ \hatf $, $ \barI_{\barhats} $,  $ \barhats $, $ X^{m} $, $ d $,$\gamma$
	\STATE $ I\gets \cup \barI_{\barhats}, \
	\lemerge\gets 
	\del_{I}(\hatf) -\thr_{\barhats,\gamma}$
	\IF{$ |\barI|=1 $}
	\STATE {\bfseries Return}: $ \lemerge $
	\ELSE
	\STATE {\bfseries Return}: $\max\{ \lemerge,\MERGE(\hatf, \barI_{\barhats_1},
 2\barhats_{1} ,  X_{1}^{m} ,d,\gamma/\sqrt{2} )$
	\STATE $ \phantom{\max\{ v,}+\MERGE(\hatf,\barI_{\barhats_2},
 2\barhats_{2} ,  X^{m}_{2} ,d, \gamma/\sqrt{2})\}$
	\ENDIF 
\end{algorithmic}
\end{algorithm}

\subsection{Distributed Computation 
of $\MERGE$ and $\STITCH$}
We consider the scenario where we are provided 
with pre-sorted samples, a known $ t $ and 
$\Theta(m)$ memory for some $t \le m \le n$. 
Let a unit of memory be equivalent to 
that which is required to store the value of one sample.
In this case, we may split the available 
memory to simulate $m$ concurrent processors 
with constant processing memory.
WLOG let $0\le t\le n$ and for simplicity,
let $t,m $ be a power of $2$, just like $n$.
Define $D_t\ed \log_2t, D_m\ed \log_2 m $ and recall
that $D\ed \log_2 n$.

Let $\STITCH_t$ be the modified $\STITCH$ that halts
in $D-D_t$ steps instead of $D$. The 
corresponding $\SURF_t$ outputs the polynomial estimate
corresponding to the interval partition given by 
$\barhatp_{D-D_t}$ 
(instead of the one corresponding to
$\barhatp_{D}$ output by $\SURF$).
Let this estimate be denoted by 
$\hatf_{\SURF_t} \ed  \hatf_{\EQ,I_{\barhatp_{D-D_t}}}$
and let 
$\bar{u}_t\ed \baru{2^{D_t}}= \baru{t}$ be the 
uniform distribution on $t$ intervals.
\begin{Lemma}
\label{lem:halt}
Given samples $ X^{n-1}\sim f $, for some 
$ t<n $ that are both powers of $ 2 $, degree $ d\le 8 $ and
the threshold $ \alpha > 2$, $ \SURF_t$ 
outputs $ \hatf_{\text{SURF}_t} $ in time 
$ \cO((d^\tau+\log n)n\log n) $ such that under event 
$ \cQ_{\epsilon} $,
\begin{align*}
\lone{\hatf_{\text{SURF}_t}-f}
\le
\min_{\barp \in \Delta_{\mathrm{bin,n}},\le 
\bar{u}_t}
\sum_{I \in \barI_{\barp}} &\left(
\frac{(r_d+1) \alpha}{\alpha-2} 
\inf_{\poly \in \cP_d}\loneint{I}{\poly-f}\right. 
\\&\quad + \left.\frac{r_d (\alpha\sqrt2+\sqrt{2}-1)}{\sqrt{2}-1}
\epsilon \sqrt{(d+1)\hatp_{I}}\right),
\vspace{-.5em}
\end{align*}
where $ q_I $ is the empirical mass under interval 
$ I $, $ r_d $ is the constant in 
Theorem~\ref{thm:polyblack}.
\end{Lemma}
As argued in Theorem~\ref{thm:interhistopt},
Lemma~\ref{lem:halt} along with Lemma~\ref{lem:binarycomp} implies that $\hatf_{\SURF_t}$ is an $r_d$-factor approximation for $\cP_{t,d}$.

For $1\le i\le D$, recall that $D(i)\ed D-i$. 
In step $i$ of $\STITCH_t$, $\MERGE$ is called on 
sub-intervals $\barI_{\barhats,j}\subseteq \barI_{\barhats}$
for $j\in\{1\upto 2^{D(i)}\}$, generated by the
interval partition $\barI_{\baru{2^{D(i)}}}$.
Each $\barI_{\barhats,j}$, for $j\in\{1\upto 2^{D(i)}\}$ 
consists of $n/2^{D(i)}=2^i$ samples. 

Given presorted samples, for the steps $i$ such that 
$2^{D(i)}\le m$, each call to $\MERGE$ may be 
implemented concurrently on the $m$ processors. 
This results in a time complexity of
$\cO((d^\tau + \log n)2^{i})$ for that step,
where $\tau\in [2,2.4)$ is the matrix inversion 
constant. As $2^{D(i)}\le m$ implies
$D(i)=D-i\le \log_2 m = D_m$, or $i\ge D-D_m$,.
The total time taken by these steps is thus given by
$\sum_{i=D-D_m}^{D-D_t} \cO((d^\tau + \log n)2^{i}) =
\cO((d^\tau + \log n)2^{D-D_t})=\cO((d^\tau + \log n)n/t)$.

For steps $i$ in the range 
$1\le i<D-D_m$, $\MERGE$ may be implemented
concurrently in batches, with each batch 
consisting of $m$ sub-intervals 
among $\barI_{\barhats,1}\upto\barI_{\barhats,j}$.
As there are a total of 
$2^{D(i)}/m$ batches, and as each interval consists 
of $2^i$ samples, step $i$ takes time
$\cO((d^{\tau}+\log n)2^i) \cdot 2^{D(i)}/m$ =
$\cO((d^{\tau}+\log n)n/m)$. 
The total time taken by steps $1\le i<D-D_m$
is given by $\cO((d^{\tau}+\log n)n/m)\cdot (D-D_m) 
= \cO((d^{\tau}+\log n)n\log n/m)$. 

Thus the time complexity under distributed computation is
$\cO((d^{\tau}+\log n)n\max\{1/t,\log n/m\})$.\label{appen:algo}
\section{Additional Experiments}
\begin{figure*}
	\centering
	\subfigure[Densities $ f_1 $, $ f_2 $.]{\begin{tikzpicture}[scale=.4]
\begin{axis}[
title={Gaussian Mixture Densities},
axis lines = left,
xlabel = $x$,
ylabel = {},
]

\addplot [
domain=0:1, 
samples=100, 
color=red,
style=dashed
]
{0.3*1/(2*pi*0.01)^(0.5)*e^(-1/2*1/0.01*(x-0.4)^2)
	+0.7*1/(2*pi*0.04)^(0.5)*e^(-1/2*1/0.04*(x-0.6)^2)
};
\addlegendentry{$ f_1(x) $}

\addplot [
domain=0:1, 
samples=100, 
color=blue,
style=solid
]
{0.4*1/(2*pi*0.0025)^(0.5)*e^(-1/2*1/0.0025*(x-0.3)^2)
	+0.6*1/(2*pi*0.0225)^(0.5)*e^(-1/2*1/0.0225*(x-0.7)^2)
};
\addlegendentry{$ f_2(x) $}
\end{axis}
\end{tikzpicture}\label{aa1}}
	\hfill
	\subfigure[$ \ell_1 $ error with $ d=1 $.]{
\begin{tikzpicture}[scale=.4]
\begin{axis}[
    title={Linear $ \SURF $},
    xlabel={Number of Samples },
    xmin=0, xmax=35000,
    ymin=0, ymax=0.17,
    xtick={0,5000,10000,15000,20000,25000,30000,35000},
    yticklabels={0,0,0.05,0.10,0.15},
    legend pos=north east,
    ymajorgrids=true,
    grid style=densely dashed,
]
 
\addplot[
    color=red,
    mark=square*,
	style=dashed
    ]
    coordinates {
    (512,.1371)(1024,0.1185)(2048,0.0958)
	(4096,0.0658)(8192,0.04738)(16384,0.0327)
	(32768,0.0258)
    };
    \addlegendentry{$ f_1 $} 

\addplot[
    color=blue,
    mark=triangle*,
    ]
    coordinates {
    (512,0.1440)(1024,.1386)(2048,0.1047)
	(4096,0.0760)(8192,0.0579)(16384,0.0419)
	(32768,0.0331)
    };
    \addlegendentry{$ f_2 $} 
\end{axis}
\end{tikzpicture}\label{bb1}}
	\hfill
	\subfigure[$ \ell_1 $ error with $ d=2 $.]{
\begin{tikzpicture}[scale=.4]
\begin{axis}[
title={Quadratic $\SURF $},
xlabel={Number of Samples },
xmin=0, xmax=35000,
ymin=0, ymax=0.17,
xtick={0,5000,10000,15000,20000,25000,30000,35000},
yticklabels={0,0,0.05,0.10,0.15},
legend pos=north east,
ymajorgrids=true,
grid style=dashed,
]

\addplot[
color=red,
mark=square*,
style=densely dashed
]
coordinates {
	(512,0.1392)(1024, 0.0843 )(2048,0.0759)
	(4096,0.0663)(8192,0.0539)(16384,0.0406)
	(32768,0.0304)
};
\addlegendentry{$ f_1 $} 

\addplot[
color=blue,
mark=triangle*,
]
coordinates {
	(512,0.1576)(1024,0.1368)(2048, 0.0974)
	(4096,0.0751)(8192,0.0558)(16384, 0.0431)
	(32768, 0.0372)
};
\addlegendentry{$ f_2 $} 
\end{axis}
\end{tikzpicture}\label{cc1}}
	\hfill
	\subfigure[$ \ell_1 $ error with $ d=3 $.]{
\begin{tikzpicture}[scale=.4]
\begin{axis}[
    title={Cubic $\SURF $},
    xlabel={Number of Samples },
    xmin=0, xmax=35000,
    ymin=0, ymax=0.17,
    xtick={0,5000,10000,15000,20000,25000,30000,35000},
    yticklabels={0,0,0.05,0.10,0.15},
    legend pos=north east,
    ymajorgrids=true,
    grid style=dashed,
]
 
\addplot[
    color=red,
    mark=square*,
	style=densely dashed
    ]
    coordinates {
    (512,0.1098)(1024, 0.0835 )(2048,0.0637)
	(4096,0.0588)(8192,0.0479)(16384,0.0436)
	(32768,0.0313)
    };
    \addlegendentry{$ f_1 $} 

\addplot[
    color=blue,
    mark=triangle*,
    ]
    coordinates {
    (512,0.1594)(1024,0.0901)(2048, 0.0726)
	(4096,0.0658)(8192,0.0591)(16384,0.0481)
	(32768,0.0251)
    };
    \addlegendentry{$ f_2 $} 
\end{axis}
\end{tikzpicture}\label{dd1}}
	\caption{Evaluation of the estimate output by $ \SURF $ 
		with degrees $ d=1,2,3 $, $ \alpha=0.25$, on
		$ f_1 = 0.3\cN( 0.4, 0.1^{2}) + 0.7\cN( 0.6, 0.2^{2})  $ and
		$ f_{2}= 0.4\cN( 0.3,0.05^{2})  + 0.6\cN( 0.7, 0.15^{2})  $. }
	\label{plot:gausssurf}
	\vspace{-1em}
\end{figure*}
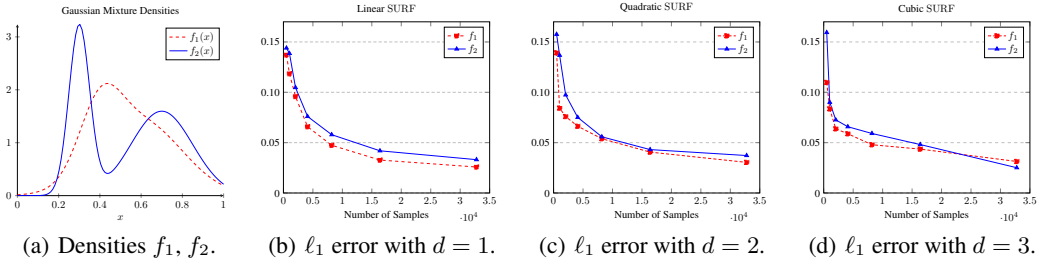
\begin{figure*}
	\centering
	\subfigure[Densities $ f_1 $, $ f_2 $.]{\begin{tikzpicture}[scale=.4]
\begin{axis}[
title={Gamma Mixture Densities},
axis lines = left,
xlabel = $x$,
ylabel = {},
]

\addplot [
domain=0:1, 
samples=100, 
color=red,
style=dashed
]
{0.2*1/(6*(.04)^4)*x^3*e^(-x/0.04)
	+0.8*1/(5040*(.06)^8)*x^7*e^(-x/0.06)
};
\addlegendentry{$ f_1(x) $}

\addplot [
domain=0:1, 
samples=100, 
color=blue,
style=solid
]
{0.4*1/(2*(.05)^3)*x^2*e^(-x/0.05)
	+0.6*1/(120*(.075)^6)*x^5*e^(-x/0.075)
};
\addlegendentry{$ f_2(x) $}
\end{axis}
\end{tikzpicture}\label{aa2}}
	\hfill
	\subfigure[$ \ell_1 $ error with $ d=1 $.]{
\begin{tikzpicture}[scale=.4]
\begin{axis}[
title={Linear $\SURF $},
xlabel={Number of Samples },
xmin=0, xmax=10000,
ymin=0, ymax=0.17,
xtick={0,2000,4000,6000,8000,10000},
yticklabels={0,0,0.05,0.10,0.15},
legend pos=north east,
ymajorgrids=true,
grid style=dashed,
]

\addplot[
    color=red,
    mark=square*,
	style=dashed
    ]
    coordinates {
    (512,0.1559)(1024,0.1192)(2048,0.0901)
	(4096,0.0642)(8192,0.0458)
    };
    \addlegendentry{$ f_1 $} 

\addplot[
    color=blue,
    mark=triangle*,
    ]
    coordinates {
    (512,0.1755)(1024,.1109)(2048,0.0782)
	(4096,0.0556)(8192,0.0466)
    };
    \addlegendentry{$ f_2 $} 
\end{axis}
\end{tikzpicture}\label{bb2}}
	\hfill
	\subfigure[$ \ell_1 $ error with $ d=2 $.]{
\begin{tikzpicture}[scale=.4]
\begin{axis}[
title={Quadratic $\SURF $},
xlabel={Number of Samples },
xmin=0, xmax=10000,
ymin=0, ymax=0.17,
xtick={0,2000,4000,6000,8000,10000},
yticklabels={0,0,0.05,0.10,0.15},
legend pos=north east,
ymajorgrids=true,
grid style=dashed,
]

\addplot[
color=red,
mark=square*,
style=densely dashed
]
coordinates {
	(512,0.1487)(1024, 0.1020 )(2048,0.0844)
	(4096,0.0689)(8192,0.0603)
};
\addlegendentry{$ f_1 $} 

\addplot[
color=blue,
mark=triangle*,
]
coordinates {
	(512,0.1121)(1024,0.0786)(2048, 0.0734)
	(4096,0.0680)(8192,0.0580)
};
\addlegendentry{$ f_2 $} 
\end{axis}
\end{tikzpicture}\label{cc2}}
	\hfill
	\subfigure[$ \ell_1 $ error with $ d=3 $.]{
\begin{tikzpicture}[scale=.4]
\begin{axis}[
title={Cubic $\SURF $},
xlabel={Number of Samples },
xmin=0, xmax=10000,
ymin=0, ymax=0.17,
xtick={0,2000,4000,6000,8000,10000},
yticklabels={0,0,0.05,0.10,0.15},
legend pos=north east,
ymajorgrids=true,
grid style=dashed,
]
 
\addplot[
    color=red,
    mark=square*,
	style=densely dashed
    ]
    coordinates {
    (512,0.1422)(1024, 0.1268 )(2048,0.0838)
	(4096,0.0751)(8192,0.0588)
    };
    \addlegendentry{$ f_1 $} 

\addplot[
    color=blue,
    mark=triangle*,
    ]
    coordinates {
    (512,0.1099)(1024,0.1045)(2048, 0.0801)
	(4096,0.0646)(8192,0.0537)
    };
    \addlegendentry{$ f_2 $} 
\end{axis}
\end{tikzpicture}\label{dd2}}
	\caption{Evaluation of the estimate output by $ \SURF $ 
		with degrees $ d=1,2,3 $, $ \alpha=0.25$, on
		$ f_1 = 0.2\text{Gam}(4,0.04) + 0.8\text{Gam}( 8, .06)  $ and
		$ f_{2}= 0.4\text{Gam}(3,0.05)  + 0.6\text{Gam}(6,.075 )  $. }
	\label{plot:gammasurf}
	\vspace{-1em}
\end{figure*}
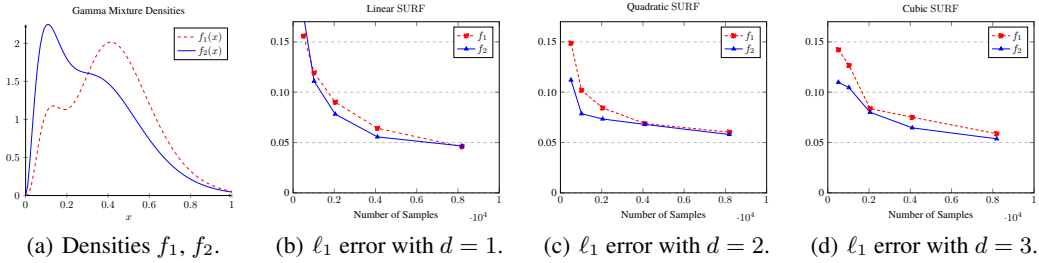
This section shows additional experiments on 
the Gaussian and gamma mixtures. 
Just as in Section~\ref{sec:experi}, 
$\SURF$ is run with $ \alpha=0.25 $ and
the results are averaged over 10 runs.

Since $\SURF$ is invariant to location-scale transformations,
WLOG we run experiments on distributions such 
that essentially all its mass lies in the interval $[0,1]$.
Let $ \cN(\mu,\sigma^2)$ be the Gaussian distribution
with parameters $ \mu $, $ \sigma $.
We run $ \SURF $ with degrees $ d=1,2,3 $ 
on the two Gaussian mixtures shown in Figure~\ref{aa1}.
Figures~\ref{bb1}--\ref{dd1} show the 
resulting $ \ell_1 $ errors. This is repeated for the gamma 
mixture density shown in Figure~\ref{plot:gammasurf}, where
$ \text{Gam}(k,\theta)$ denotes the gamma distribution
with shape, scale parameters $ k$, $ \theta $ respectively.
Figures~\ref{bb2}--\ref{dd2} show the corresponding
$\ell_{1}$ errors.

Notice that the
errors are similar between distributions,
and that the error saturates more quickly 
for $d=3$, as the higher degree allows $\SURF$ to exploit the
smoothness inherent in the considered parametric families.
These observations are in line with what was observed for
the beta mixtures considered in Figure~\ref{plot:betasurf}.\label{appen:experi}
\end{document}